%% file: main_new.tex
\documentclass[lettersize,journal]{IEEEtran}
\usepackage{booktabs}
\usepackage{multirow}
\usepackage{tabularx}
\usepackage{arydshln}
\usepackage[utf8]{inputenc}
\usepackage[strings]{underscore}
\usepackage{threeparttable}
\usepackage{amsmath,amsfonts}
\usepackage{enumitem}
\usepackage{algorithmic}
\usepackage{algorithm}
\usepackage{array}
\usepackage[caption=false,font=normalsize,labelfont=sf,textfont=sf]{subfig}
\usepackage{textcomp}
\usepackage{stfloats}
\usepackage{url}
\usepackage{verbatim}
\usepackage{graphicx}
\usepackage{authblk}
\usepackage[breaklinks=true]{hyperref}
\usepackage[hyphens]{xurl}
\usepackage[normalem]{ulem}
\usepackage[xcolor,commandnameprefix=ifneeded,final]{changes}
\setaddedmarkup{\textcolor{blue}{\uline{#1}}}
\setdeletedmarkup{\textcolor{red}{\sout{#1}}}

\usepackage[acronym]{glossaries}
\newacronym{ai}{AI}{artificial intelligence}
\newacronym{mesh}{MeSH}{medical subject headings}
\newacronym{pci}{PCI}{percutaneous coronary intervention}
\newacronym{pvi}{PVI}{peripheral vascular intervention}
\newacronym{dsa}{DSA}{digital subtraction angiography}
\newacronym{ivus}{IVUS}{intravascular ultrasound}
\newacronym{ep}{EP}{electro physiological}
\newacronym{rcms}{RCMS}{robotic catheter manipulation system}
\newacronym{ip camera}{IP camera}{internet protocol camera}
\newacronym{rvir}{RVIR}{remote-controlled vascular interventional robot}
\newacronym{fbg}{FBG}{Fiber Bragg Grating}
\newacronym{smlr}{SMLR}{sparse multivariate linear regression}
\newacronym{vpn}{VPN}{virtual private network}
\newacronym{trl}{TRL}{Technology Readiness Levels}
\newacronym{fda}{FDA}{Food and Drug Administration}
\newacronym{lan}{LAN}{local area network}
\newacronym{isac}{ISAC}{Integrated Sensing and Communication}
\newacronym{mt}{MT}{mechanical thrombectomy}
\newacronym{vir2}{VIR-2}{vascular intervention robot 2}
\newacronym{ar}{AR}{augmented-reality}
\newacronym{or}{OR}{operating room}

\usepackage{siunitx}
\usepackage{cite}
\usepackage{color}
\usepackage{xcolor}
\newenvironment{DIFnomarkup}{}{}
\begin{document}
\bstctlcite{IEEEexample:BSTcontrol}
\title{Remote Teleoperation of Endovascular Intervention Robots: A Systematic Review}

\author{Xingyu Chen, Yinchao Yang, Nikola Fischer, Harry Robertshaw, Benjamin Jackson, Mohammad Shikh-Bahaei, Christos Bergeles, Thomas C Booth 
        % <-this % stops a space
% \thanks{$^{1}$Surgical and Interventional Engineering, School of Biomedical Engineering \& Imaging Sciences, King's College London, London, UK. $^{2}$Department of Engineering, King's College London, UK. $^{3}$ Department of Neuroradiology, King's College Hospital, London, UK.}%<-this % stops a space
\thanks{Xingyu Chen, Nikola Fischer, Harry Robertshaw, Benjamin Jackson, Christos Bergeles are with School of Biomedical Engineering \& Imaging Sciences, King's College London, WC2R 2LS London, U.K.}%
\thanks{Yinchao Yang and Mohammad Shikh-Bahaei are from Department of Engineering, King's College London, WC2R 2LS London, U.K.}%
\thanks{Thomas C Booth is with School of Biomedical Engineering \& Imaging Sciences, King's College London, WC2R 2LS London, U.K., and also with Department of Neuroradiology, King's College Hospital, SE5 9RS London, U.K. (email: thomas.booth@kcl.ac.uk)}%
\thanks{The work of Xingyu Chen was supported by the King’s College London–China Scholarship Council
Joint Ph.D.\ Scholarship (K-CSC)}%
\thanks{This work was supported by the MRC IAA 2021 Kings College London (MR/X502923/1) and the Wellcome EPSRC Centre for Medical Engineering at King’s College London (203148/Z/16/Z)}%
}%



\maketitle

\begin{abstract}
Remote robotic-assisted endovascular intervention offers a promising approach to reduce clinician radiation exposure and physical strain, while extending specialized vascular care to geographically distant regions. Despite advancements, teleoperated endovascular intervention remains underexplored, especially for time-sensitive interventions like mechanical thrombectomy for acute stroke. 
The aim of the current review was to determine the evidence regarding teleoperated endovascular robotic systems, covering technical feasibility, communication infrastructure, and clinical outcomes. The review further identified research gaps and future directions.
Following PRISMA guidelines, 16 studies were included that met the inclusion criteria out of 2501 initial search results. We found that teleoperated catheters and guidewires, driven by mechanical or electromagnetic systems, can be navigated across distances up to \SI{7000}{km}. With robust communication infrastructure, network latency remained within clinically acceptable limits (\SI{30}{}-\SI{163}{\milli\second}). Although initial outcomes highlighted 100\% procedural success in small-scale human trials, most evidence stemmed from animal or phantom models.
Overall, the findings suggest that teleoperated endovascular intervention can reduce occupational hazards, expand patient access to urgent procedures, and optimize resource allocation.
Future research should be conducted in low and middle income countries to demonstrate broader geographical access. Ultimately, multi-center clinical trials are required to validate the safety, efficacy, and generalization in diverse clinical settings.

\end{abstract}


\section{Introduction} \label{intro}
\input{Sections/Introduction}

\section{Method} \label{Method}

\input{Sections/Method}

\section{Results} \label{Results}
\input{Sections/Results}

\section{Discussion} \label{Discussion}
\input{Sections/Discussion}

\input{main_new.bbl}

\end{document}

%% file: Sections/Introduction.tex
% 1. (Interventional radiology,Minimally invasive vascular interventional surgery, fluoroscopy and muscle strain)
\IEEEPARstart{E}{ndovascular} techniques have rapidly evolved in interventional radiology and cardiology, allowing physicians to perform minimally invasive treatment with increased precision and reduced patient morbidity~\cite{miller_overview_2008}.  
Nonetheless, navigating catheters and devices through tortuous vascular anatomies is challenging because it requires high levels of dexterity, concentration, and endurance~\cite{rafii-tari_current_2014}.
It is also noteworthy that prolonged exposure to radiation places clinicians at risk of harm, while the physical demands of wearing heavy protective gear to mitigate the exposure contribute to musculoskeletal strain and occupational hazards, including long-term orthopedic issues~\cite{miller_occupational_2010,ross_prevalence_1997}. 

% 2. (To protect operator, robotic systems for endovascular intervention,)
%To address the health risks and ergonomic challenges associated with conventional endovascular procedures, researchers and clinicians have explored the use of robotic systems specifically designed for endovascular interventions~\cite{crinn   ion2022robotics}. These endovascular robots were invented with the primary goal of protecting operators from harmful radiation while simultaneously reducing physical fatigue~\cite{madder_impact_2017}. Remote control of catheters and guide wires through robotic systems can improve procedural precision and diminish operator errors, ultimately improving overall safety and effectiveness~\cite{riga_role_2010}. 
%Furthermore, integrating \gls{ai} has the potential to amplify these benefits~\cite{robertshaw_artificial_2023,feizi2021robotics} ([8,9]). Moreover, their precise control and assistance capabilities have the potential to standardize complex interventions and facilitate advanced techniques. Community and rural surgeons are also increasingly trained in laparoscopy and robotic operations, gaining access to advanced endovascular platforms that could make teleoperated interventions more widely attainable across diverse practice settings~\cite{schneider2021inequalities}({10}).
To address the health risks and ergonomic challenges associated with endovascular procedures, engineers and clinicians investigated the use of robotic systems specifically developed for these interventions~\cite{crinnion2022robotics}. It was noted that some endovascular robots are developed with the primary goal of protecting operators from harmful radiation exposure while simultaneously reducing physical fatigue~\cite{madder_impact_2017}. Additional incentives for teleoperated robot development include improving procedural precision of catheters and guidewires and reducing operator error, which likely improves clinical safety and effectiveness~\cite{riga_role_2010}.

% 3. (Different types of robotic system,  catheter and soft continuum robot, magnetic actuated robot, commercial systems, including different components of the robot)
%The development of endovascular robots has undergone several phases, integrating diverse engineering principles to address distinct clinical needs. Among the prominent designs are catheter robots that enable precise positioning of endovascular instruments~\cite{wei2024telemanipulated}, magnetic actuated robots leveraging external magnetic fields for navigation~\cite{von_arx_simultaneous_2024,kim2022telerobotic}, and continuum robots that offer flexible, snake-like movement within intricate vascular pathways~\cite{rafii-tari_current_2014,bao_multilevel_2022}. In recent years, commercial robotic systems have emerged, featuring user-friendly interfaces~\cite{pereira_first--human_2020}, sophisticated control algorithms~\cite{nelson_remote_2024}, and improved safety measures~\cite{bell_e-069_2024,nicholls_mark_2021}. These advancements continue to progress robotic solutions towards widespread clinical adoption.
The mechatronic development of endovascular robots has been multidirectional, integrating diverse engineering principles to address distinct clinical needs. Among the prominent designs are catheter robots that enable precise positioning of endovascular instruments~\cite{wei2024telemanipulated}, magnetic-actuated robots leveraging external magnetic fields for navigation~\cite{von_arx_simultaneous_2024,kim2022telerobotic}, and continuum robots that offer flexible, snake-like movements~\cite{rafii-tari_current_2014,bao_multilevel_2022}. In parallel, commercial robotic systems have emerged, featuring user-friendly interfaces~\cite{pereira_first--human_2020}, advanced control algorithms~\cite{nelson_remote_2024}, and a focus on safety measures~\cite{bell_e-069_2024,nicholls_mark_2021}. More recently, the integration of \gls{ai} has been proposed to amplify safety and effectiveness benefits~\cite{robertshaw_artificial_2023}, because precise control and assistance capabilities might facilitate complex interventions.

These technical advancements continue to progress robotic solutions towards the clinic. Furthermore, widespread translation of endovascular robots is also becoming more feasible due to an increase in endovascular operators in both specialist and non-specialist hospitals globally, as well as an increased understanding of robotics amongst clinicians~\cite{schneider2021inequalities}.

\begin{figure}[!t]
\centering
\includegraphics[width=1\linewidth]{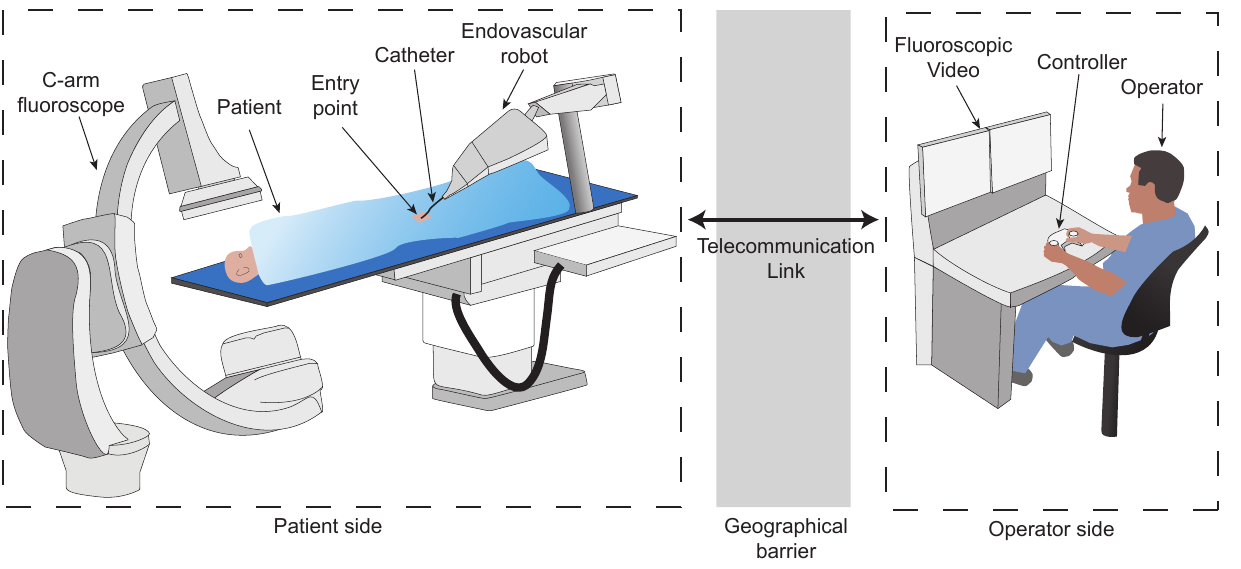}
\caption{Illustration of a teleoperated endovascular intervention setup. The endovascular robot is positioned at the patient side to manipulate the catheter under real-time fluoroscopic imaging. Control commands and fluoroscopic video streams are transmitted bidirectionally through a telecommunication link between the patient side and the remote operator side, enabling the operator to perform the procedure across geographical barriers.}
\label{fig:endo_illu}
\end{figure}

Despite notable progress in robotic endovascular interventions, the process of teleoperation remains insufficiently addressed in the literature. In the context of endovascular robotics, teleoperation denotes a telerobotic architecture in which a patient side robot is physically located in the \gls{or}, while the operator controls its actuators in real time from a remote master console via a communication network. This configuration, including the spatial separation between the operator and patient, and the bidirectional exchange of control commands and fluoroscopic imaging, is illustrated in Figure~\ref{fig:endo_illu}.
Remote teleoperated endovascular robots have the potential to expand patient access to urgent procedures, such as \gls{mt} for acute ischemic stroke~\cite{fransen_time_2016}, by enabling expert operators to perform interventions from geographically distant locations~\cite{mcmeekin_estimating_2017}. \gls{mt} is time sensitive, with previous studies showing that treatment under a six hour window can effectively decrease disability, and for every hour of delay the treatment benefit will be dramatically reduced~\cite{goyal_endovascular_2016}. In the United Kingdom, for instance, only 3.9\% of eligible stroke patients receive MT~\cite{Healthcare2024stroke}, primarily due to the scarcity of expert operators and the patchy geographic distribution of specialized centers~\cite{saver_time_2016}. In comparison, 8.4\% of ischemic-stroke patients in Germany receive \gls{mt}~\cite{ungerer2024evolution}; in the United States, the figure is 5.6\% across all 50 states~\cite{jafarli2025trends}; and in China the national endovascular therapy rate for acute ischemic stroke was just 1.45\%~\cite{ye2022rates}. These statistics underscore that restricted access to \gls{mt} is a global issue.

An effective teleoperative capability could enable remote robotic \gls{mt} and optimize resource allocation~\cite{picozzi2023telemedicine}. By reducing treatment delays~\cite{sacks_multisociety_2018}, mitigating specialist shortages, and extending rapid access to urgent interventions, such capability would plausibly improve clinical outcomes. 
Although endovascular robotic systems are currently available, some platforms support telerobotic intervention in which the clinician operates from a nearby control room; however, these local-area configurations do not introduce additional wide-area telecommunication delays, and therefore do not reflect the performance challenges associated with long-distance teleoperation~\cite{eleid_remote_2021,madder_network_2020}. Consequently, many existing systems still lack a reliable and proven remote teleoperative capability because of telecommunication constrains~\cite{barba2022remote}, global network constraints, high financial costs, and regulatory requirements~\cite{picozzi2023telemedicine}. Moreover, not all procedural steps can currently be performed by robot, and the majority of existing studies therefore focus primarily on robotic navigation rather than full procedure automation. Device preparation, vascular access, contrast injection, clot retrieval and instrument exchange still commonly require a trained on-site team, owing to limitations in available robotic hardware, incomplete automation of tool manipulation, and safety requirements. However, endovascular navigation has been identified as one of the stages requiring the highest level of operator expertise~\cite{lenthall2017bsnr}. Enabling this navigation task to be performed remotely would substantially reduce the expertise required from on-site personnel. 

Remote teleoperated surgery (telesurgery) has been investigated in various clinical scenarios with the transatlantic Lindbergh operation in 2001 being one of its earliest instances~\cite{ marescaux_transatlantic_2001}. However, seamless interaction between remote operators and robotic systems remains underdeveloped, particularly in scenarios requiring both real-time operation and high precision. To mitigate this, modern teleoperated robotic (telerobotic) platforms increasingly depend on advanced telecommunication infrastructure. For instance, 5G networks can reduce telecommunication latency to imperceptible levels~\cite{legeza_impact_2022} and achieve consistent performance without interruptions~\cite{moustris_long_2023}. These innovations underpin critical functionalities such as real-time visual feedback~\cite{choi_telesurgery_2018} and haptic control~\cite{yuan_teleoperation_2021}. Additional telecommunication methods like \gls{isac}~\cite{liu2020joint}, semantic communication~\cite{luo2022semantic}, and edge intelligence~\cite{luo2021resource} also have potential to further boost telesurgery performance. The implementation of encrypted protocols ensures the safeguarding of patient data~\cite{xu_determination_2014,madder_network_2020}. Concurrently, dedicated surgical communication frameworks designed for uninterrupted operator-robot data exchange are being refined to ensure reliability under clinical demands~\cite{li_design_2021}. 

% Dedicated surgical communication framework for real-time data exchange between operators and robotic systems to achieve consistent performance without interruptions~\cite{li_design_2021} and solve the latency control~\cite{legeza_impact_2022} and cybersecurity has be developed based on the development of telecommunication. \cite{xu_determination_2014,madder_network_2020}
% Continue here

%Despite progress,  persistent challenges threaten broader adoption. Vulnerabilities in operator-robot communication, such as  data packet loss can threat procedural safety. Ethical concerns, including regulatory inconsistencies and risks of cyberattacks targeting sensitive medical data, further hinder widespread adoption~\cite{butner_transforming_2003}. A systematic evaluation of current research is therefore essential to advance solutions that enhance the security, responsiveness, and clinical viability of remote endovascular systems. This review aims to provide a robust evidence base, offering insights into clinical best practices, technological advancements, and future directions for the teleoperation of endovascular robot research.
Despite progress in telerobotics, persistent challenges threaten broader adoption. Vulnerabilities in operator-robot communication, such as data packet loss can threaten procedural safety. Ethical concerns, regulatory challenges, and risks of cyberattacks targeting sensitive medical data, further hinder widespread adoption ~\cite{butner_transforming_2003}.

Given the increasing interest in teleoperated endovascular robots and the challenges surrounding teleoperation, it is essential to summaries the current landscape. This paper systematically evaluates the current evidence relating to the teleoperation of endovascular robots. Specifically, it provides a robust evidence base, and offers insights into clinical practices, technological advancements, and future directions for the teleoperation of endovascular robot research.

% 7. (main contribution of this paper)
% This systematic review will employ the Preferred Reporting Items for Systematic Reviews and Meta-Analyses (PRISMA) guidelines to ensure a comprehensive and transparent evaluation of available evidence~\cite{page_prisma_2021}.  Inclusion criteria focus on peer-reviewed studies involving both preclinical and clinical applications of teleoperated robotic systems. Data extraction will emphasize safety, efficacy, and feasibility outcomes, while rigorous quality assessment tools will be applied to minimize bias. By adopting this structured approach, the review aims to provide a robust evidence base, offering insights into clinical best practices, technological advancements, and future directions for teleoperated endovascular robot research. Following this introduction, the review is structured into methods, results, and discussion sections, culminating in recommendations for future research and clinical implementation.

%% file: Sections/Method.tex
\subsection{Selection Criteria}
\begin{figure*}[!t]
\centering
\includegraphics[width=0.9\linewidth]{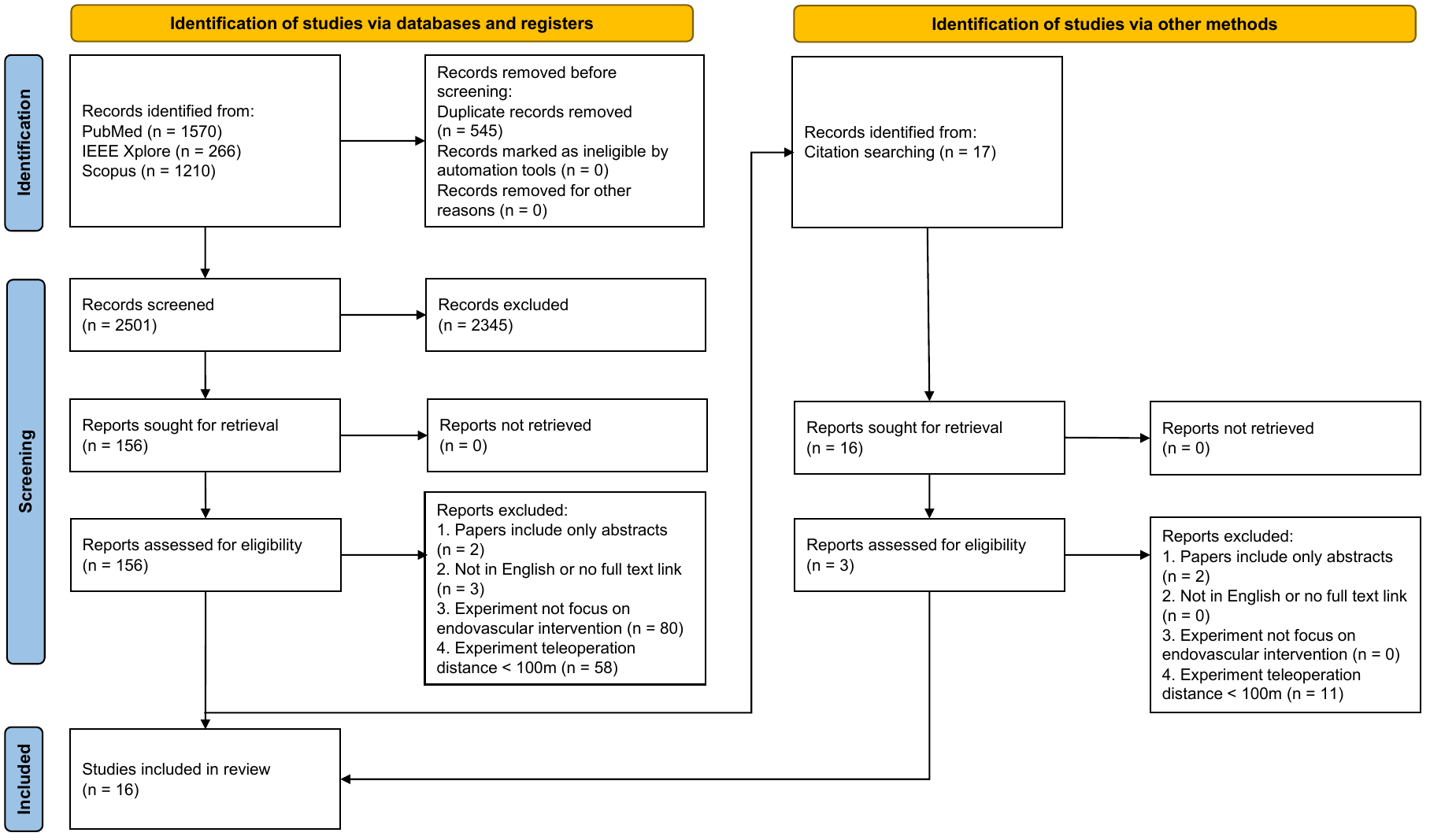}
\caption{PRISMA flow diagram showing the number of articles searched and excluded at each stage of the literature search after screening titles, abstracts, and full texts.}
\label{fig:PARISMA_flowchart}
\end{figure*}

We performed a systematic review by identifying and synthesizing evidence on teleoperation in endovascular robotic interventions using the Preferred Reporting Items for Systematic Reviews and Meta-Analyses (PRISMA) guidelines~\cite{page_prisma_2021-1}.  

\subsubsection{Information Sources and Search Strategy}
A search was conducted in three electronic databases: PubMed, IEEE Xplore, and Scopus up to March 19, 2025.  The search was conducted using the keywords and \gls{mesh} terms: (“remote OR teleoperate OR telerobotic OR telestenting OR telesurgery OR telemanipulate OR telecommunications”) AND (“endovascular OR vascular OR neurovascular OR neuroendovascular OR percutaneous coronary OR cardiovascular OR catheter OR guidewire”) AND (“intervention OR surgery OR navigation OR interventional”). The searching keywords can be found in supplementary table 1. 

% A detailed combinatorial presentation of the search terms is provided in Table~\ref{tab:search_terms}.  
% %%%%MESH Table Here
% \begin{table}[!ht]
% \centering
% \caption{Systematic review search terms}
% \begin{tabular}{ccc}
% %\hline
% \textbf{TELEOPERATION} & \textbf{ENDOVASCULAR} & \textbf{INTERVENTION} \\ \hline
% remote                          & endovascular                   & intervention                   \\ \hline
% teleoperate                     & vascular                       & surgery                        \\ \hline
% telerobotic                     & neuroendovascular              & navigation                     \\ \hline
% telestenting                    & percutaneous coronary          & interventional                 \\ \hline
% telesurgery                     & cardiovascular                 &                                \\ \hline
% telemanipulate                  & catheter                       &                                \\ \hline
% telecommunications              & guidewire                      &                                \\ \hline
%                                 & neurovascular                  &                                \\ \hline
% \end{tabular}
% \label{tab:search_terms}
% \end{table}
% %%%%%%

\subsubsection{Eligibility Criteria}
Studies eligible for inclusion were original research articles investigating teleoperation in endovascular robotic interventions, specifically highlighting procedures where the primary catheter or guidewire navigation task was performed remotely. Studies that did not involve teleoperated manipulation of endovascular instruments, such as those limited to robot kinematics, tip-tracking performance, or benchtop accuracy assessments unrelated to procedural tasks were excluded.   
Due to the main review focus on remote teleoperation, those studies involving teleoperated interventions conducted at a distance of less than \SI{100}{\meter} were excluded.
% , unless a clearly specified telecommunication link and accompanying communication performance metrics were reported, demonstrating their relevance to wider-area or remote scenarios.
Articles without full-text availability in English were also excluded. Review articles, technical briefs, abstracts lacking full-text versions, or other non-original research contributions were excluded. To incorporate the most recent technological and clinical developments, only studies published in or after 2010 were considered.  
 
\subsubsection{Selection and Data Collection Process}
All identified records initially underwent title and abstract screening to remove clearly irrelevant or ineligible items based on the eligibility criteria. Studies deemed potentially eligible were retrieved in full text for thorough assessment. Data extracted from the included studies is presented in subsequent sections. 

\subsection{Data Items, Effect Measures, and Synthesis Methods}
The following data elements were extracted from studies meeting the eligibility criteria: study characteristics (authors, publication year, journal or conference), clinical setting (e.g., type of surgical procedure, \textit{in vivo}, \textit{in vitro}), robotic platform type, teleoperation distance, communication technologies utilized, and the number of experimental procedures. Telecommunication parameters such as latency, security measures, and video streaming specifications were also recorded. Reported outcomes included procedural success rates, procedure duration, and any noted ergonomic benefits or limitations.

\subsection{Study Risk of Bias and Certainty Assessment} 
\label{M-risk}
To evaluate methodological rigor and potential biases in the included studies, this systematic review employed two established tools: the Quality Assessment of Diagnostic Accuracy Studies 2 (QUADAS-2)~\cite{whiting_quadas-2_2011} and the Risk Of Bias In Non-randomized Studies of Interventions (ROBINS-I)~\cite{sterne_robins-i_2016}. These frameworks were used to guide the structured appraisal of study design, implementation, and reporting quality. 

\subsection{Meta-analysis}
We were unable to perform a meta-analysis due to insufficient homogenous studies. 

%% file: Sections/Results.tex
%\subsection{Overview of teleoperation progress}
% write as order of diffrent robots
% Studies Table showing common teleoperation characteristics( distance, latency, etc)
% Commercial robot systems achieved teleoperation.
\input{Sections/tables/table_trial}
A total of 2501 studies were included after the initial search, with 156 remaining after title and abstract screening, and 16 (\!\!\cite{guo_internet_2012,xiao_internet-based_2012,lu_experimental_2013,patel_long_2019,madder_network_2020,madder_feasibility_2019,zhang_improved_2021,legeza_preclinical_2021,singer_remote_2021,eleid_remote_2021,madder_robotic_2021,yang_cloud_2022,serafini_exploring_2022,han_multi-device_2023,xu_long-distance_2024,madder_transatlantic_2025}) meeting the inclusion criteria after full-text screening. Fig.~\ref{fig:PARISMA_flowchart} illustrates the selection process, while Table~\ref{tab:general} summarizes key characteristics. The included studies predominantly focused on \gls{pci} (8/16, 50\%), followed by general endovascular interventions (6/16, 38\%), with single studies addressing cardiac ablation and \gls{mt}. 

Experimental stages varied across studies: 8/16 (50\%) used \textit{in vitro} phantom models~\cite{guo_internet_2012,xiao_internet-based_2012,legeza_preclinical_2021,singer_remote_2021,madder_robotic_2021,yang_cloud_2022,serafini_exploring_2022,madder_transatlantic_2025}, 6/16 (38\%) performed \textit{in vivo} experiments in animals~\cite{lu_experimental_2013,madder_feasibility_2019,madder_network_2020,zhang_improved_2021,eleid_remote_2021,han_multi-device_2023}, and 2/16 (12.5\%) studies were clinical trials in \textit{humans}~\cite{patel_long_2019,xu_long-distance_2024}.

During experimental trials, each system typically involved two procedural stages: 1) On-Site Team Operation Stage and 2) Robot Operation Stage. During the On-Site Team Operation Stage, the local clinical staff handled tasks that could not be performed remotely. These tasks encompassed the initial insertion of the catheter, the replacement of devices, the injection of contrast agent, and the verification of the final placement of devices as directed by the remote operator. The on-site team was additionally responsible for maintaining the patient’s stability and any urgent procedures. During the Robot Operation Stage, tasks were generally confined to guiding the catheter or guidewire to its target location. Specifically, 7/16 (44\%) studies report that the remote operator controls only catheter or guidewire navigation~\cite{guo_internet_2012,xiao_internet-based_2012,lu_experimental_2013,madder_network_2020,zhang_improved_2021,yang_cloud_2022,serafini_exploring_2022}, whereas in 9/16 (56\%) other studies, the remote operator also managed additional device placement tasks such as placing a stent or inflating a balloon~\cite{patel_long_2019,legeza_preclinical_2021,singer_remote_2021,eleid_remote_2021,madder_robotic_2021,han_multi-device_2023,xu_long-distance_2024,madder_transatlantic_2025,madder_feasibility_2019}. Current teleoperation capabilities demonstrated considerable geographical reach, with a maximum reported distance of \SI{9000}{\kilo\meter} for transatlantic procedures~\cite{nelson_remote_2024}. Clinical applications showed promising outcomes, with 100\% technical success rates in both clinical trials~\cite{patel_long_2019,xu_long-distance_2024}. Table~\ref{tab:general} shows the key parameters of each study including: operational distance, surgical procedure, the robotic platform used, and the exact tasks assigned to the robot-operation and the on-site team.

Risk of bias assessment using ROBINS-I and QUADAS-2 tools revealed methodological limitations. All studies exhibited high bias risks due to imprecise or absent patient selection criteria, inconsistently described reference standards, and minimal definition of specific teleoperated endovascular intervention protocols used in each trial. Inconsistent or undefined patient groups, reference standards and index tests reduce the reliability and comparability of reported outcomes including index test performance. Details of the risk-of-bias assessment findings are available in the supplementary Table 2.  Despite these constraints, the low quality evidence serves as baseline data which will be useful when designing future studies.

\begin{DIFnomarkup}
\begin{table*}[!t]
\centering
\caption{Description of different communication methods \added{used in the included studies}}
\label{tab:des_com}
\begin{tabularx}{\textwidth}{@{}cX@{}}
% \toprule
\multicolumn{1}{c}{\textbf{Communication Method}} & 
\multicolumn{1}{c}{\textbf{Description}} \\
\midrule
\textbf{Internet \added{(unspecified access)}} & 
Refers to a general purpose, public network connection where data travel over various network tiers. \added{This category is applied when a study only stated that the connection was “via the internet” with no specification of communication protocols, techniques, or mechanisms involved, and instead only} provided a broad reference to internet-based connectivity.\\
\midrule
\textbf{5G} & 
The fifth-generation mobile network standard offering high bandwidth, low latency, and improved reliability. \added{Within this classification, 5G denotes the use of a mobile network as the primary wireless access interface before traffic is forwarded to external networks such as the public Internet or private 5G slices.}\\
\midrule
\textbf{Ethernet} & 
\added{A wired local area network technology that provides very high throughput, ultra-low latency, and high reliability, but offers no mobility. In this classification, ethernet refers to wired access within buildings or institutional networks that} offers reliable, high-speed data transmission with low latency \added{before data is forwarded to wider network infrastructures.} \\
\midrule
\textbf{Wired} & 
A direct point-to-point physical connection using dedicated fiber optics. \added{Studies categorized under this label explicitly describe the use of dedicated fiber circuits, which provide a} highly controlled transmission environment with minimal electromagnetic interference, delivering predictable low latency and exceptional data integrity over long distances. \\

\midrule
\textbf{Wi-Fi} & 
A wireless networking technology based on IEEE 802.11 standards, \added{offering high throughput and low latency within a limited coverage area, with moderate reliability and limited mobility. In this classification, wi-fi refers to local wireless access through a router}, which then routes data to the broader internet infrastructure.\\

% \bottomrule
\end{tabularx}
\end{table*}
\end{DIFnomarkup}

\subsection{Robotic Platforms}
\begin{figure}[!ht]
\centering
\includegraphics[width=1\linewidth]{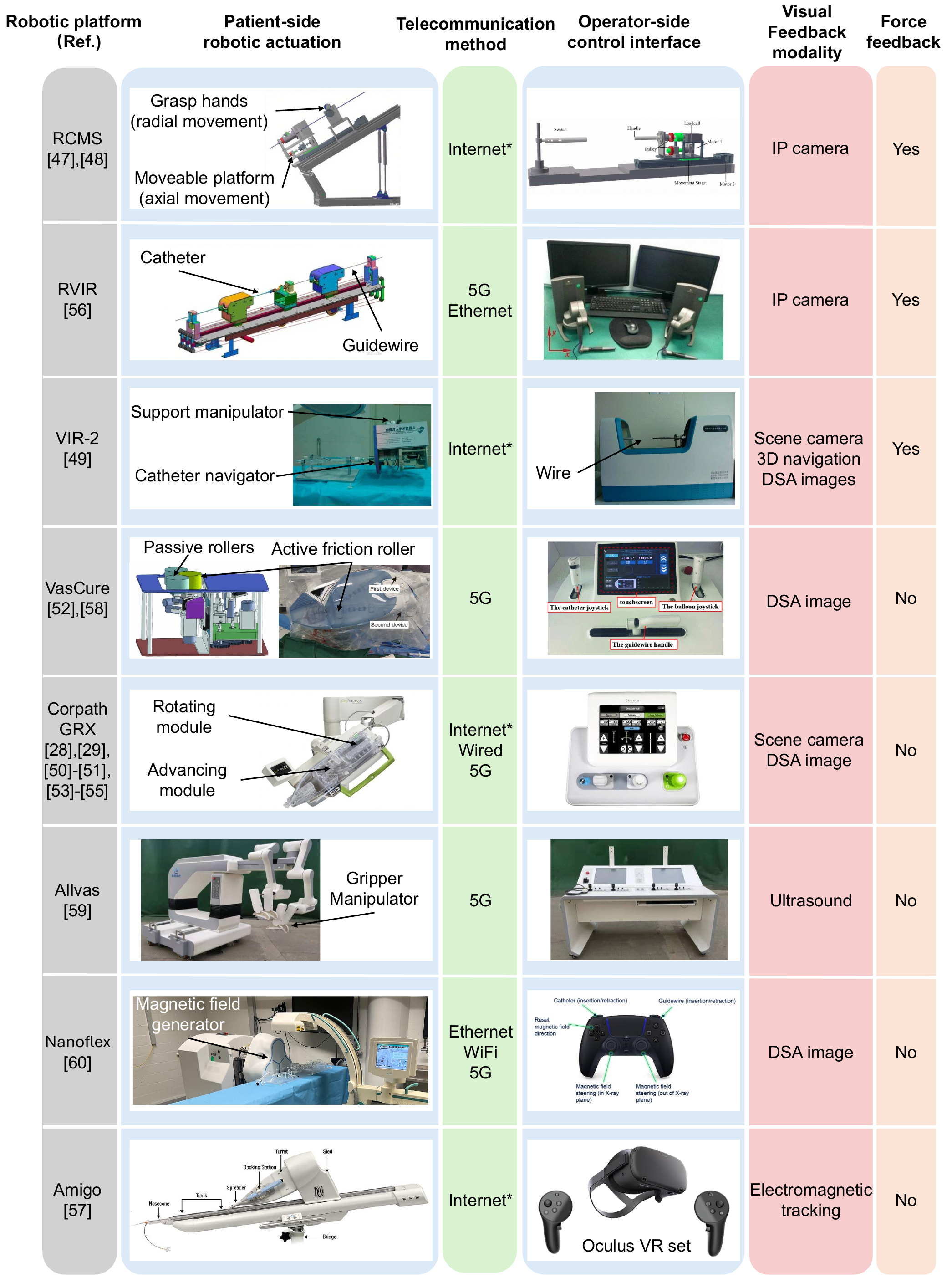}
\caption{Comparative analysis of robotic surgical platforms and their interfaces. Including their communication methods and force/visual feedback mechanisms. \\ *Internet communication denotes platforms where authors did not specify detailed communication methods but broadly referenced internet-based connectivity. \\RCMS: robotic catheter manipulation system, RVIR: remote-controlled vascular interventional robot, VIR-2: vascular intervention robot 2, IP camera: internet protocol camera, DSA: digital subtraction angiography.}
\label{fig:robotic_platforms}
\end{figure}

Various robotic platforms were utilized across the studies reviewed. Specifically, 6/16 (38\%) studies employed experimental robotic platforms while 10/16 (63\%) studies used commercial robotic platforms. Eight different robotic platforms are utilized in the included studies, with four (50\%) in the experimental stage and four (50\%) as commercial robotic systems. Fig.~\ref{fig:robotic_platforms} provides a systematic comparison of key robotic platforms and their teleoperated interfaces, communication methods, and force/visual feedback mechanisms.

Regarding the classification of robotic platforms based on interventional therapy applications, two primary categories emerged: general vascular interventional robots and \gls{ep} interventional therapy robots. These two types of interventional procedures exhibit significant differences in terms of treatment approaches, objectives, and required instrumentation, which subsequently influence the design of the respective robots. Among the studies reviewed, 15/16 (94\%) focused on general vascular intervention, while one study~\cite{serafini_exploring_2022} addressed \gls{ep} therapy (discussed later in this section). General vascular interventional robots are primarily designed for procedures such as angioplasty, intravascular infusion, and clot removal, whereas \gls{ep} therapy robots are specifically aimed at ablation procedures.

Beyond the previously assessed metrics, the instrument handling capabilities for each robotic platform was further characterized. Among the eight platforms, 6/8 (75\%) operated standard passive instruments (commercial catheters, microcatheters, and guidewires), whereas two platforms supported active instruments (with active instruments defined as those equipped with one or more steerable segments). Navion~\cite{madder_transatlantic_2025} employed an active catheter with a magnetized distal segment, while the Amigo platform~\cite{serafini_exploring_2022} enabled the control of a steerable ablation catheter. For each platform, the number of endovascular instruments controlled simultaneously (e.g., guidewire only, catheter and guidewire, or multi-device configurations) was also documented. Across eight platforms, 3/8 (38\%) were capable of manipulating only a single instrument, while another 3 (38\%) enabled the control of two instruments simultaneously, and 2 (25\%) reported the ability to control three or more instruments concurrently. Regarding robotic functions demonstrated in anatomical settings, as summarized in Table~\ref{tab:general}, 4/8 (50\%) platforms were limited to basic catheter and guidewire navigation, while 3/8 (38\%) platforms demonstrated additional procedural capabilities such as balloon angioplasty and stent deployment. One platform, Amigo, further reported support for advanced electrophysiological tasks, including electroanatomic mapping and cardiac ablation. 

%%%% Experiment stage robor systems
Experimental platforms were used in six studies (6/16, 38\%). In terms of actuation methods, only one robot platform used a magnetic field generator to operate the robot, whereas all other systems relied on mechanical actuation, i.e., direct manipulation of the instrument by mechanical force and torque transmission. Additionally, only 3/8 (38\% in eight robotic platforms) robotic platforms included haptic feedback capabilities. Visual feedback was obtained via direct camera feeds in two systems (25\%), \gls{dsa} images in four systems (50\%), ultrasound in one system (12.5\%), and electromagnetic tracking in one system (12.5\%). The following sections provide a comprehensive examination of each robotic platform.
%Nanoflex
\paragraph{\textbf{Navion}}
An electromagnetic navigation system (Navion, Nanoflex Robotics AG, Opfikon, Switzerland) ~\cite{madder_transatlantic_2025} successfully performed transatlantic \gls{mt} over a distance exceeding \SI{9000}{\kilo\meter}~\cite{nelson_remote_2024}. 
The Navion robotic system operates through magnetic field actuation, whereby a mobile magnetic field generator directs controlled magnetic fields towards the patient, enabling the manipulation of a magnetized guidewire tip for precise navigation to the target anatomy. In the reported teleoperation configuration, the Navion system was controlled from a remote console connected via several communication pathways, including: 1) WebSocket over the public Internet 2)Ethernet and WIFI, 3)WIFI and 5G, and 4)WIFI alone. Operator feedback consisted exclusively of streamed DSA images, as no haptic feedback was available. The platform supported manipulation of a single active, magnetically actuated guidewire, whereas device loading and contrast injection was performed manually by the on-site team.  

\paragraph{\textbf{RCMS/RVIR}}
A \gls{rcms}, incorporated a linear slide mechanism for device delivery with integrated force feedback capabilities~\cite{guo_internet_2012,xiao_internet-based_2012}. This system included controller-responder interaction enabled by visual feedback from an \gls{ip camera}. The system was later upgraded to the \gls{rvir} platform~\cite{bao_cooperation_2018}, which maintained \gls{rcms}' core actuation principles while introducing dual-instrument operation and enhanced force feedback via \gls{fbg} sensors, while also providing visual feedback from an \gls{ip camera}. The robot was further extended by enabling the use of cloud server~\cite{yang_cloud_2022}. In its original configuration, \gls{rcms} controls a single passive instrument, while the \gls{rvir} expanded this capability to coordinated dual passive control of both catheter and guidewire. Both robotic platforms did not report any tasks beyond instrument navigation.   

\paragraph{\textbf{VIR-2}}
The \gls{vir2} platform integrated dual-angle \gls{dsa} with 3D vascular reconstruction for precise visual positioning, complemented by a catheter-mounted fiber optic pressure sensor for force feedback~\cite{lu_experimental_2013,lu_application_2011}, with data transmitted through Internet connection. While the actuation mechanism was not fully detailed, it reportedly combined a supporting robotic arm paired with a catheter navigator, featuring a dedicated user interface so that operators could handle a wire in a manner that simulates actual catheter manipulation. \gls{vir2} was able to control only a single passive instrument for navigation, while all other procedural steps were performed manually.  

\paragraph{\textbf{VasCure}}
5G-enabled PCI was demonstrated using the VasCure robotic platform~\cite{zhang_improved_2021,zhao_design_2021}. This system employed one active and two passive rollers for device manipulation, with axial rotation facilitated by a subjacent lifting mechanism. It was reported to support simultaneous manipulation of two passive instruments and to execute balloon angioplasty and stent placement. VasCure lacked integrated force feedback, however, the authors highlighted plans for future torque sensor integration. This platform’s capabilities were further validated through dedicated 5G network trials transmitting both control signals and \gls{dsa} images~\cite{han_multi-device_2023}.

%% commercial systems
\paragraph{\textbf{Corpath GRX}}
%Corpath GRX
Seven studies (44\%) reported clinical use of the CorPath GRX system (Corindus Vascular Robotics, Inc., now Siemens Healthineers, Newton, USA) for teleoperated procedures~\cite{patel_long_2019,madder_feasibility_2019,madder_network_2020,legeza_preclinical_2021,singer_remote_2021,eleid_remote_2021,madder_robotic_2021}. CorPath GRX uses motor-driven friction wheels to advance and rotate instruments, and, across these studies, could sequentially manipulate up to three passive instruments. Its remote variant enables communication via public Internet, dedicated wired fiber links, or 5G networks between operator consoles and robotic units, with DSA images streamed to the operator and no haptic feedback available. The system further demonstrated the ability to perform remote stent deployment and balloon angioplasty. 

\paragraph{\textbf{Allvas}}
%Allvas
The Allvas robotic system (Aopeng Medical Technology Co., Ltd, Shanghai, China) was used to demonstrate long-distance procedures~\cite{xu_long-distance_2024}. The robot was actuated with dual robotic arms and gripper-embedded manipulators to allow clamping and movement of instruments~\cite{song_novel_2023}, and was guided by \gls{ivus}. Each gripper manipulator handles a passive catheter or guidewire, and the system can coordinate up to four instruments to support navigation, balloon angioplasty, and stent deployment. In the included studies, the remote console and the robot were connected via a commercial 5G network, with intravascular ultrasound transmitted to the operator and no dedicated haptic feedback reported.

%%% ablation
\paragraph{\textbf{Amigo}}
%Amigo
Interventional \gls{ep} therapy consists of treating cardiac arrhythmias using specialized tools like radiofrequency ablation catheters. These devices deliver targeted thermal energy to cardiac tissue to correct abnormal electrical pathways. The Amigo robotic system (Catheter Precision Inc., Fort Mill, USA) was shown to be capable of transcontinental \gls{ep} procedures~\cite{serafini_exploring_2022}. The Amigo system provides motorized manipulation of a single active steerable ablation catheter and is able to perform electroanatomic mapping and cardiac ablation remotely. In the study~\cite{serafini_exploring_2022}, the leader controller and follower robot were connected via a general Internet-based communication link, with electroanatomical mapping data transmitted and no haptic feedback reported.  

Teleoperated endovascular interventions have successfully been demonstrated using other commercial robotic systems outside of the peer-reviewed literature. Whilst not included for systematic analysis, we provide a brief mention of these systems to allow a more comprehensive understanding of advancements in the field:

\begin{itemize}
    %Xcath
    \item XCath System: An endovascular robotic system developed by XCath (XCath Inc., Houston, USA) implemented live broadcast teleoperations using electronically-controlled steerable guidewires during international medical conferences~\cite{bell_xcath_2024,bell_xcath_2024-1}. Their electro-steerable guidewire robotic system was operated remotely from Abu Dhabi on a phantom model located in South Korea at a distance of \SI{6900}{\kilo\meter}. It supports mechanically driven advancement of an electronically controlled active steerable guidewire; in the remote trial, fluoroscopic video was transmitted via a commercial internet connection with no haptic feedback reported. The system was also reported to enable positioning and deployment of thrombectomy devices.   
    %Remedy robotics
    \item Remedy Robotics: Remedy Robotics (Remedy Robotics Inc., San Francisco, USA) successfully performed a remote endovascular navigation in a silicone model over a distance of \SI{12800}{\kilo\meter}, from Melbourne, Australia, to Toronto, Canada~\cite{bell_e-069_2024}. While no specific details of its actuation method were provided, Remedy Robotics reported the ability to control up to four passive instruments simultaneously, and no additional procedural capabilities were described. In the experiments, DSA video streams were transmitted via commercial internet and VPN connections, and no haptic feedback was available.  
    %Robocath
    \item R-One System: The Robocath R-One robotic system (Robocath SAS, Rouen, France) was used to conduct remote endovascular intervention experiments. Progressing from \SI{120}{\kilo\meter} in animal trials (2021)~\cite{nicholls_mark_2021} to \SI{2800}{\kilo\meter} in human coronary interventions (2023)~\cite{noauthor_robocath_nodate}. The R-one robotic systems uses friction wheels to control a single passive instrument and was reported to support stent placement and balloon angioplasty. In the remote experiments, DSA image streams were transmitted via a 5G connection, and no haptic feedback was available.   

\end{itemize}

\subsection{Telecommunication Framework}
\begin{figure*}[!t]
\centering
\includegraphics[width=1\linewidth]{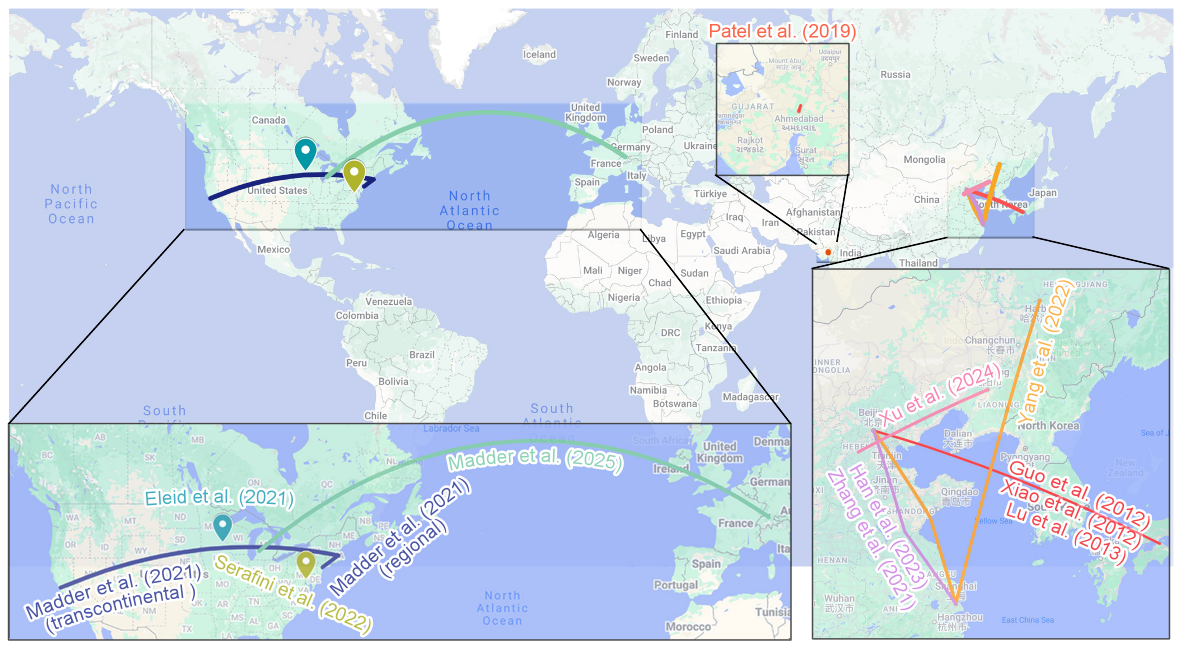}
\caption{An overview of the global layout of teleoperated endovascular intervention experiments, with each line illustrating the distinct city-to-city connections that have been tested in each study. The figure is created using Google maps.}
\label{fig:worldMap}
\end{figure*}
%%% put a common circle of the transmission with data type and size on it. 
% characteristics table of telecommunication part

% communication method
The systematic review found that most studies lacked detail regarding the telecommunication technologies used, with most systems (12/16, 75\%) using general commercial internet connections (Table~\ref{tab:des_com})~\cite{guo_internet_2012,xiao_internet-based_2012,lu_experimental_2013,patel_long_2019,madder_feasibility_2019,madder_network_2020,legeza_preclinical_2021,singer_remote_2021,eleid_remote_2021,madder_robotic_2021,serafini_exploring_2022,madder_transatlantic_2025}. 
Four studies (4/16, 25\%) implemented common commercial 5G networks (see Table~\ref{tab:des_com} for description)~\cite{zhang_improved_2021,yang_cloud_2022,han_multi-device_2023,xu_long-distance_2024}, including one investigation employing a dedicated 5G line for enhanced reliability~\cite{han_multi-device_2023}. 
Two studies~\cite{madder_network_2020,madder_robotic_2021} additionally incorporated dedicated wired connections into their experimental communication system to evaluate data transmission reliability during testing. 

Geographically, most experiments took place in regions with established telecommunication infrastructures such as the USA and China (Fig.~\ref{fig:worldMap}), with a paucity of teleoperation studies in low and middle income countries. 
Although remote experiments may involve geographically dispersed participants, the majority of current research remains classified as single-center due to their centralized coordination by a single institution. 

% communication video stream
Bidirectional communication architectures between on-site teams and remote operators were implemented in 10/16 (63\%) studies. Among these, half the studies (5/10) employed pre-existing video conference software, namely Jitsi  (Jitsi.org)~\cite{madder_network_2020}, LifeSize (Lifesize Inc., Austin, USA)~\cite{legeza_preclinical_2021,madder_impact_2017}, Voov meeting (Tencent Inc., Hong Kong, China)~\cite{yang_cloud_2022} and Zoom (Zoom Inc., San Jose, USA)~\cite{serafini_exploring_2022}. The remaining five studies~\cite{patel_long_2019,madder_feasibility_2019,singer_remote_2021,eleid_remote_2021,xu_long-distance_2024} developed specialized multi-camera systems with dedicated video streaming methods to facilitate procedural guidance.

% operation images
Three distinct strategies to track wires and catheters emerged from the analysis: 5/16 (31\%) studies~\cite{guo_internet_2012,xiao_internet-based_2012,lu_experimental_2013,singer_remote_2021,yang_cloud_2022} employed conventional optical cameras, 7/16 (44\%) studies~\cite{patel_long_2019,madder_feasibility_2019,madder_network_2020,zhang_improved_2021,legeza_preclinical_2021,eleid_remote_2021,madder_robotic_2021} utilized real-time fluoroscopic or angiographic imaging, while one study~\cite{serafini_exploring_2022} implemented a simulation-based guidance system. There, catheter position was monitored with a tracking system (Aurora, Northern Digital Inc, Shelburne, USA) and subsequently rendered incorporating with the phantom in a virtual three-dimensional display for the remote operator.

% latency reported, packet loss rate
Latency is the time delay between data transmission and reception. It is a critical telecommunication parameter in teleoperated endovascular intervention procedures, where real-time device control relies on uninterrupted communication. Across 12/16 studies (75\%) studies, latency was explicitly measured, with all reported values falling below the \SI{250}{\milli\second} threshold generally considered safe for \gls{pci} procedures; this \SI{250}{\milli\second} threshold was established based on systematic evaluation of how latency affects procedural outcomes~\cite{eleid_remote_2021,madder_network_2020}.
While latency was prioritized for its direct impact on procedural safety and operator feedback, other parameters like packet loss rate, bandwidth consistency, and jitter (variation in packet delay) were rarely analyzed. These metrics, though relevant to network performance, were omitted in most studies due to technical limitations in the testing environments. 

\begin{figure}[!t]
\centering
\includegraphics[width=1\linewidth]{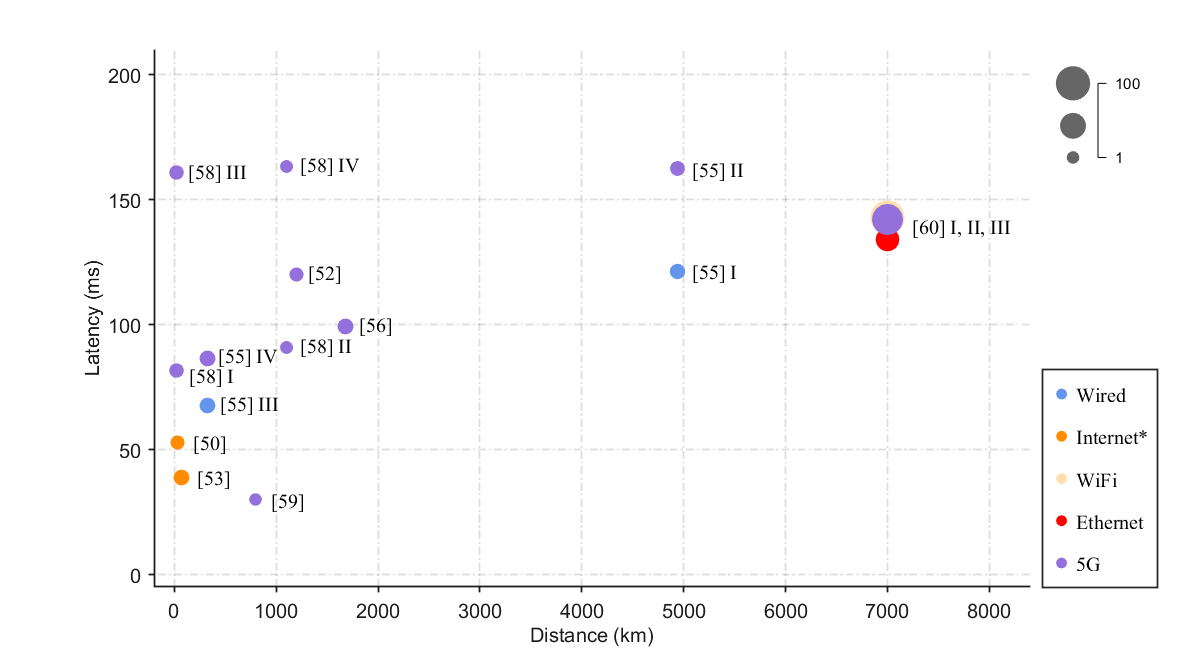}
\caption{Latency variability compared to distance under diverse experimental conditions. Each circular marker corresponds to a specific experimental setup (denoted by Roman number) within various studies. Each color represents a particular connection method, while the circular marker’s size denotes the volume of procedures conducted for each experimental setup. \\ *Platforms where authors did not specify detailed communication methods but broadly referenced internet-based connectivity.}
\label{fig:latency_visualize}
\end{figure}

A detailed visualization of the relationship between distance and latency is shown in Fig.~\ref{fig:latency_visualize}. When distance was under \SI{2000}{\kilo\meter}, reported latencies ranged between \SI{30}{}-\SI{163}{\milli\second}. In contrast, for distances exceeding \SI{4000}{\kilo\meter}, latencies spanned from \SI{121}{}-\SI{162}{\milli\second}. 
% In a study which investigated transfer of different information, command transmissions showed less delay than image transfer: \SI{90}{\milli\second} compared to \SI{163}{\milli\second} at \SI{1100}{\kilo\meter}, and \SI{81}{\milli\second} compared to \SI{161}{\milli\second} at \SI{18.3}{\kilo\meter}~\cite{han_multi-device_2023}. Furthermore, in a study comparing communication methods and teleoperation latency across three distinct network configurations (5G, Ethernet and WiFi), no notable latency difference was observed (\SI{134}{}–\SI{143}{\milli\second}). In terms of managing latency, one study used a \gls{smlr} model to predict and mitigate latency in teleoperated systems, achieving prediction errors below \SI{4.5}{\milli\second} under fluctuating network loads and abrupt delay spikes~\cite{yang_cloud_2022}.

In one study that investigated transfer of different information, image transfer showed higher delays than command transmissions, rising from \SI{90}{\milli\second} to \SI{163}{\milli\second} at \SI{1100}{\kilo\meter} distance and from \SI{81}{\milli\second} to \SI{161}{\milli\second} at \SI{18.3}{\kilo\meter} distance~\cite{han_multi-device_2023}. This suggests that the size of the transmitted data exerts a substantially greater influence than the variation in transmission distance. Another study revealed no notable connection between communication methods and teleoperation latency, as experiments across three distinct network configurations (5G, Ethernet, and Wi-Fi) yielded comparable latency outcomes (\SI{134}{\milli\second}–\SI{143}{\milli\second})~\cite{madder_transatlantic_2025}. In Han \textit{et al.}’s work \cite{han_multi-device_2023}, tests I and II represented latency for transport command signals, whereas experiments III and IV targeted image transmission. The image transfer showed higher delays than command transmissions, rising from \SI{90}{\milli\second} to \SI{163}{\milli\second} at \SI{1100}{\kilo\meter} and from \SI{81}{\milli\second} to \SI{161}{\milli\second} at \SI{18.3}{\kilo\meter}. However, latency variations are notably influenced by heterogeneous study parameters such as network load and procedural complexity, which differ across studies. These results are further constrained by limited experimental repetitions, preventing definitive conclusions about optimal communication protocols and operational distance. 

In terms of managing latency, one study used a sparse multivariate linear regression (SMLR) model to predict and mitigate latency in teleoperated systems, achieving prediction errors of network latency below \SI{4.5}{\milli\second} under fluctuating network loads and abrupt delay spikes~\cite{yang_cloud_2022}.

Only 4/16 (25\%) studies implemented \gls{vpn} or firewalls to secure their data transmission. One study employed a standard \gls{vpn} during communication~\cite{madder_feasibility_2019}, while three studies utilized both \gls{vpn} and hardware firewalls simultaneously. Among these three, one study lacked detailed information about the firewall hardware~\cite{singer_remote_2021}, whereas the remaining two~\cite{legeza_preclinical_2021,madder_robotic_2021} both using FortiGate (Fortinet Inc., Sunnyvale, USA) firewall to safeguard network data. Additional quantitative metrics about telecommunication technologies used for every teleoperation experiment reviewed are collated in Supplementary Table 3.

% Figure showing the latency versus distance(communication method, transnational).

\subsection{Clinical Evaluation}
\input{Sections/tables/Clinical_table_3}
% Figures showing their target regions
\begin{figure}[!t]
\centering
\includegraphics[width=1\linewidth]{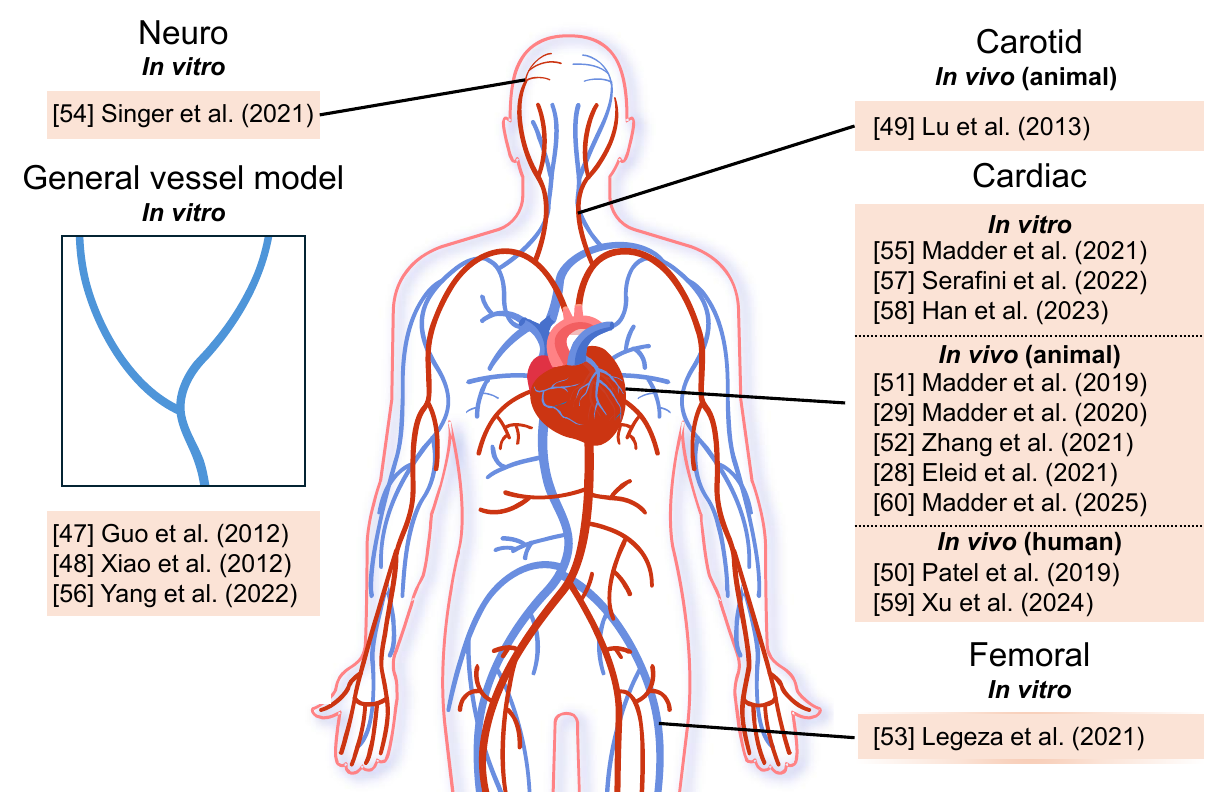}
\caption{Diagram depicting the anatomical region of interest for each study.}
\label{fig:vessel_region}
\end{figure}

Across all studies, most individual experiments were performed using \textit{in vitro} phantoms ($n= 152–260$ ~\cite{madder_network_2020,madder_transatlantic_2025}). Fewer \textit{in vivo} ($n = 22$ from ~\cite{lu_experimental_2013,madder_feasibility_2019,zhang_improved_2021,eleid_remote_2021,han_multi-device_2023}) and \textit{human interventions} ($n= 6$ from ~\cite{patel_long_2019,xu_long-distance_2024}) were performed. Fig.~\ref{fig:vessel_region} illustrates the anatomical focus of the studies. Of these, 10/16 (63\%) focused on cardiac interventions, 3/16 (19\%) employed generalized vascular models without anatomical specificity, while the remaining studies addressed neurovascular ($n= 1$), carotid ($n= 1$), and femoral ($n= 1$) interventions. Nine studies (9/16, 56\%) used vascular phantoms, including: the EVE silicone simulator (FAIN-Biomedical Inc., Okayama, Japan)~\cite{guo_internet_2012,xiao_internet-based_2012,yang_cloud_2022}; the ANGIO Mentor endovascular simulator (Simbionix USA Corporation, Cleveland, USA)~\cite{legeza_preclinical_2021,madder_network_2020,madder_robotic_2021}; a fluid-filled silicone vasculature model (United Biologics Inc., Irvine, USA)~\cite{singer_remote_2021}; custom cardiac models fabricated from polymethyl methacrylate (PMMA; Plexiglas) with additively manufactured intracardiac structures~\cite{serafini_exploring_2022} (detailed design of the custom model is reported in~\cite{seslar_initial_2018}) and the Trando 3D Neuro Vascular System I (Ningbo Trando 3D Medical Technology Co., Ltd., Ningbo, China)~\cite{madder_transatlantic_2025}. One hybrid study combined swine models with phantom testing using the ANGIO Mentor endovascular simulator~\cite{madder_network_2020}. Six preclinical studies (6/16, 38\%) employed animal models: one used beagle canines~\cite{lu_experimental_2013}, five used swine~\cite{madder_feasibility_2019,madder_network_2020,zhang_improved_2021,han_multi-device_2023, eleid_remote_2021}. Clinical trials were limited to two studies (2/16, 13\%)~\cite{patel_long_2019,xu_long-distance_2024}, both of which centered on PCI and reported 100\% procedural success across six remote operator-guided interventions.

Table~\ref{tab:clinical} summarizes results of the experimental evaluation across the reviewed studies, including operational efficiency metrics and latency impacts. \textit{Human trials} demonstrated robust success rates, with procedural success in all cases for PCI at latencies of \SI{53}{\milli\second}~\cite{patel_long_2019} and \SI{30}{\milli\second}~\cite{xu_long-distance_2024}, respectively. Although latencies below \SI{250}{\milli\second} were generally imperceptible (\!\!\cite{madder_network_2020};\cite{eleid_remote_2021}), a noticeable delay was experienced below this commonly-accepted threshold in one study~\cite{yang_cloud_2022}.

Operational duration varied substantially, with fluoroscopy times ranging from $6.5 \pm 1.8$\,\SI{}{\minute} during \textit{in vitro} \gls{pvi}~\cite{legeza_preclinical_2021} to \SI{104.3}{\minute} during porcine \gls{pci}~\cite{han_multi-device_2023}, attributable to variation in procedural complexity. \textit{Human trials} reported shorter procedure durations, with interventions completed in a mean time of \SI{23.6}{\minute} (range: $19-29$\,\SI{}{\minute}) ~\cite{patel_long_2019}.

%% file: Sections/tables/table_trial.tex
% \usepackage{booktabs}
% \usepackage{graphicx}
\begin{table*}[!t]
\centering
\caption{Studies included from the PRISMA search strategy. Information included: working distance, procedure type, robotic platform, robot and on-site perform stage.}
\label{tab:general}
\resizebox{2\columnwidth}{!}{%
\begin{tabular}{@{}lcccccc@{}}
%\toprule
\multicolumn{1}{c}{Study} &
  Operating distance (km) &
  Type of surgical procedure &
  Cases (n) &
  Robotic platform &
  Robot operation stage &
  On-site team operation stage \\ \midrule
\begin{tabular}[c]{@{}l@{}}\cite{guo_internet_2012} Guo \textit{et al.} (2012)\\ \cite{xiao_internet-based_2012} Xiao \textit{et al.} (2012)\end{tabular} &
  1679 &
  EI &
  \begin{tabular}[c]{@{}c@{}}2\\ 7\end{tabular} &
  RCMS &
  Catheter navigation &
  Initial vascular access \\ \midrule
\cite{lu_experimental_2013} Lu \textit{et al.} (2013) &
  2500 &
  CA &
  1 &
  VIR-2 &
  Catheter navigation &
  Initial vascular access \\ \midrule
\cite{patel_long_2019} Patel \textit{et al.} (2019) &
  32 &
  PCI &
  5 &
  CorPath GRX &
  \begin{tabular}[c]{@{}c@{}}Guidewire/ Catheter navigation\\ Stent implantation\end{tabular} &
  \begin{tabular}[c]{@{}c@{}}Initial vascular access\\ Device loading\end{tabular} \\ \midrule
\cite{madder_feasibility_2019} Madder \textit{et al.} (2019) &
  \begin{tabular}[c]{@{}c@{}}6.6 \\ 166 \end{tabular} &
  PCI &
  \begin{tabular}[c]{@{}c@{}}2\\ 3\end{tabular} &
  CorPath GRX &
  \begin{tabular}[c]{@{}c@{}}Guidewire/ Catheter navigation\\ Stent implantation\end{tabular} &
  Initial vascular access \\ \midrule
\cite{madder_network_2020} Madder \textit{et al.} (2020) &
  \begin{tabular}[c]{@{}c@{}}7.36 (\textit{in vitro})\\ 6.6 (\textit{in vivo} 1)\\ 165 (\textit{in vivo} 2)\end{tabular} &
  PCI &
  \begin{tabular}[c]{@{}c@{}}95 (\textit{in vitro})\\ 57 (\textit{in vivo})\end{tabular} &
  CorPath GRX &
  Guidewire/ Catheter navigation &
  \begin{tabular}[c]{@{}c@{}}Initial vascular access\\ Guidewire positioning\\ Device loading\end{tabular} \\ \midrule
\cite{zhang_improved_2021} Zhang \textit{et al.} (2021) &
  1200 &
  PCI &
  5 &
  VasCure &
  Guidewire/ Catheter navigation &
  \begin{tabular}[c]{@{}c@{}}Initial vascular access\\ Device loading\\ Any required manual interventions\end{tabular} \\ \midrule
\cite{legeza_preclinical_2021} Legeza \textit{et al.} (2021) &
  70.4 &
  PVI &
  10 &
  CorPath GRX &
  \begin{tabular}[c]{@{}c@{}}Guidewire/ Catheter navigation\\ Stent implantation\\ Balloon angioplasty\end{tabular} &
  \begin{tabular}[c]{@{}c@{}}Initial vascular access to the arterial\\ Device exchanges\end{tabular} \\ \midrule
\cite{singer_remote_2021} Singer \textit{et al.} (2021) &
  8 &
  MT &
  — &
  CorPath GRX &
  \begin{tabular}[c]{@{}c@{}}Guidewire/ Catheternavigation\\ Thrombectomy device positioning\end{tabular} &
  \begin{tabular}[c]{@{}c@{}}Initial vascular access\\ Aspiration and retrieval of thrombus\end{tabular} \\ \midrule
\cite{eleid_remote_2021} Eleid \textit{et al.} (2021) &
  1.6 &
  PCI &
  8 &
  CorPath GRX &
  \begin{tabular}[c]{@{}c@{}}Guidewire/ Catheter navigation\\ Balloon angioplasty\\ Stent implantation\end{tabular} &
  \begin{tabular}[c]{@{}c@{}}Initial vascular access to the ascending aorta\\ Device loading\end{tabular} \\ \midrule
\cite{madder_robotic_2021} Madder \textit{et al.} (2021) &
  \begin{tabular}[c]{@{}c@{}}330 (regional)\\ 4936 (transcontinental)\end{tabular} &
  PCI &
  \begin{tabular}[c]{@{}c@{}}20\\ 16\end{tabular} &
  CorPath GRX &
  \begin{tabular}[c]{@{}c@{}}Guidewire/ Catheter navigation\\ Stent implantation\end{tabular} &
  \begin{tabular}[c]{@{}c@{}}Initial vascular access to the ascending aorta\\ Device loading\end{tabular} \\ \midrule
\cite{yang_cloud_2022} Yang \textit{et al.} (2022) &
  1676 &
  EI &
  10 &
  RVIR &
  Guidewire/ Catheter navigation &
  \begin{tabular}[c]{@{}c@{}}Initial vascular access\\ Device loading\end{tabular} \\ \midrule
\cite{serafini_exploring_2022} Serafini \textit{et al.} (2022) &
  3380 &
  CCA &
  — &
  Amigo &
  Guidewire/ Catheter navigation &
  Initial vascular access \\ \midrule
\cite{han_multi-device_2023} Han \textit{et al.} (2023) &
  \begin{tabular}[c]{@{}c@{}}18.3 \\ 1100 \end{tabular} &
  PCI &
 \begin{tabular}[c]{@{}c@{}}6\\ 2\end{tabular} &
  VasCure &
  \begin{tabular}[c]{@{}c@{}}Guidewire/ Catheter navigation\\ Stent implantation\\ Balloon angioplasty\end{tabular} &
  \begin{tabular}[c]{@{}c@{}}Initial angiography\\ Initial vascular access\\ Device loading\end{tabular} \\ \midrule
\cite{xu_long-distance_2024} Xu \textit{et al.} (2024) &
  800 &
  PCI &
  1 &
  ALLVAS robotics &
  \begin{tabular}[c]{@{}c@{}}Guidewire/ Catheter navigation\\ Stent implantation\end{tabular} &
  \begin{tabular}[c]{@{}c@{}}Initial vascular access\\ Device loading\end{tabular} \\ \midrule
\cite{madder_transatlantic_2025} Madder \textit{et al.} (2025) &
  7000 &
  ICA &
  260 &
  Nanoflex Robotics &
  Guidewire/ Catheter navigation &
  \begin{tabular}[c]{@{}c@{}}Initial vascular access\\ Device loading\\ Contrast Injection\end{tabular} \\ \midrule
\multicolumn{6}{c}{PVI: Peripheral vascular intervention, MT: Mechanical thrombectomy, EI: Endovascular intervention, CA: Cerebral angiography, CCA: Cardiac catheter ablation, ICA: Invasive coronary angiography} \\ %\bottomrule
\end{tabular}%
}
\end{table*}

%% file: Sections/tables/Clinical_table_3.tex
% \usepackage{booktabs}
% \usepackage{multirow}
% \usepackage{graphicx}
\begin{table*}[!t]
\centering
\caption{Summary of Clinical Outcomes Including Success Rates, Operational Efficiency, and Latency Impacts}
\label{tab:clinical}
\resizebox{2\columnwidth}{!}{%
\begin{tabular}{@{}lccccccc@{}}
% \toprule
\multicolumn{1}{c}{\multirow{2}{*}{Study}} &
  \multirow{2}{*}{Procedure} &
  \multirow{2}{*}{Model} &
  \multicolumn{2}{c}{Participants} &
  \multirow{2}{*}{Operation Time} &
  \multirow{2}{*}{Latency} &
  \multirow{2}{*}{Operator Evaluation} \\ \cmidrule(lr){4-5}
\multicolumn{1}{c}{} &
   &
   &
  Cases (n) &
  Surgeons (n) &
   &
   &
   \\ \midrule
\cite{guo_internet_2012} Guo \textit{et al.} (2012) &
  EI &
  In vitro &
  2 &
  — &
  — &
  — &
  — \\ \midrule
\cite{xiao_internet-based_2012} Xiao \textit{et al.} (2012) &
  EI &
  In vitro &
  7 &
  — &
  — &
  300 ms &
  — \\ \midrule
\cite{lu_experimental_2013} Lu \textit{et al.} (2013) &
  CA &
  In vivo (animal) &
  1 &
  1 &
  45 min &
  1 s &
  — \\ \midrule
\cite{patel_long_2019} Patel \textit{et al.} (2019) &
  PCI &
  In vivo (human) &
  5 &
  5 &
  23.6 (19–29) min &
  53 ms &
  \begin{tabular}[c]{@{}c@{}}Latency imperceptible\\ Control and communication satisfactory\end{tabular} \\ \midrule
\cite{madder_feasibility_2019} Madder \textit{et al.} (2019) &
  PCI &
  In vivo (animal) &
  5 &
  — &
  — &
  — &
  — \\ \midrule
\cite{madder_network_2020} Madder \textit{et al.} (2020) &
  PCI &
  \begin{tabular}[c]{@{}c@{}}In vitro\\ In vivo (animal)\end{tabular} &
  152 &
  — &
  \begin{tabular}[c]{@{}c@{}} 33.0 secs\\ 17.9 secs\end{tabular} &
  150–1000 ms &
  \begin{tabular}[c]{@{}c@{}}Latency \textless{} 250 ms: Latency imperceptible\\ Latency \textgreater{} 250 ms: significantly perceived\end{tabular} \\ \midrule
\cite{zhang_improved_2021} Zhang \textit{et al.} (2021) &
  PCI &
  In vivo (animal) &
  5 &
  — &
  — &
  120 ms &
  — \\ \midrule
\cite{legeza_preclinical_2021} Legeza \textit{et al.} (2021) &
  PVI &
  In vitro &
  10 &
  — &
    6.5 ± 1.8 min &
  38.9 ms &
  Communication satisfactory \\ \midrule
\cite{singer_remote_2021} Singer \textit{et al.} (2021) &
  MT &
  In vitro &
  — &
  1 &
  — &
  — &
  — \\ \midrule
\cite{eleid_remote_2021} Eleid \textit{et al.} (2021) &
  PCI &
  In vivo (animal) &
  8 &
  3 &
  5.1 ± 3.3 min&
  150–1000 ms &
  Latency \textgreater{} 250 ms: perceived by operator \\ \midrule
\multirow{2}{*}{\cite{madder_robotic_2021} Madder \textit{et al.} (2021)} &
  \multirow{2}{*}{PCI} &
  \multirow{2}{*}{In vitro} &
  \multirow{2}{*}{36} &
  \multirow{2}{*}{1} &
  \begin{tabular}[c]{@{}c@{}}R*, Wired: 9.0 min\\ R, 5G: 6.3 min\end{tabular} &
  \begin{tabular}[c]{@{}c@{}}R, Wired: 67.8 ms\\ R, 5G: 86.6 ms\end{tabular} &
  \multirow{2}{*}{Latency rated as imperceptible} \\ 
  % \cmidrule(lr){6-7}
 &
   &
   &
   &
   &
  \begin{tabular}[c]{@{}c@{}}T*, Wired: 4.1 min\\ T, 5G: 3.0 min\end{tabular} &
  \begin{tabular}[c]{@{}c@{}}T, Wired:  121.5 ms\\ T, 5G: 162.5 ms\end{tabular} &
   \\ \midrule
\cite{yang_cloud_2022} Yang \textit{et al.} (2022) &
  EI &
  In vitro &
  10 &
  — &
  — &
  99.1 ms &
  Able to sense the latency even latency \textless{} 250 ms \\ \midrule
\cite{serafini_exploring_2022} Serafini \textit{et al.} (2022) &
  CCA &
  In vitro &
  — &
  — &
  — &
  — &
  Latency imperceptible \\ \midrule
\cite{han_multi-device_2023} Han \textit{et al.} (2023) &
  PCI &
  In vivo (animal) &
  8 &
  — &
  104.3 min &
  \begin{tabular}[c]{@{}c@{}}Command: 80 - 90 ms\\ Image: 161 - 163 ms\end{tabular} &
  — \\ \midrule
\cite{xu_long-distance_2024} Xu \textit{et al.} (2024) &
  PCI &
  In vivo (human) &
  1 &
  — &
  — &
  30 ms &
  Latency imperceptible \\ \midrule
\cite{madder_transatlantic_2025} Madder \textit{et al.} (2025) &
  ICA &
  In vitro &
  260 &
  2 &
  33.2 secs &
  \begin{tabular}[c]{@{}c@{}}Ethernet: 134 ms\\ WiFi: 143 ms\\ 5G: 142 ms\end{tabular} &
  Latency had no or minor impact on procedure \\ \midrule
\multicolumn{8}{l}{PVI: Peripheral Vascular Intervention, MT: Mechical Thrombectomy, EI: Endovascular Intervention, CA: Cerebral Angiography, CCA:Cardiac Catheter Ablation, ICA: Invasive Coronary Angiography} \\ 
\multicolumn{8}{l}{R*: Regional, T*: Transcontinental}
% \bottomrule
\end{tabular}%
}
\end{table*}

%% file: Sections/Discussion.tex
\subsection{Summary of Teleoperated Endovascular Interventions}
This systematic review compared and contrasted advancements in teleoperated endovascular interventions from 16 studies, revealing both technological progress and unresolved challenges. It addressed a literature gap as the first comprehensive analysis focusing explicitly on teleoperation capabilities in endovascular robotics. 
Current evidence confirms that robotic systems integrated with advanced telecommunication infrastructure can successfully perform procedures over long distances, including \SI{7000}{\kilo\meter} transatlantic operations~\cite{madder_transatlantic_2025} and \SI{12800}{\kilo\meter} ultra-long-ranges~\cite{bell_e-069_2024}. This capability was enabled by synergistic combinations of telecommunication technologies, such as dedicated 5G networks for low-latency command signal transmission; hybrid communication frameworks (e.g., wired-wireless integration). Commercial platforms like the CorPath GRX and Allvas were used to validate the feasibility of teleoperation in clinical and experimental settings. Experimental platforms (RVIR, VasCure) have employed predictive latency models. Clinical trials for \gls{pci} achieved 100\% technical success rates despite distance of \SI{800}{\km}, highlighting the feasibility of these systems in controlled environments. Furthermore, the integration of real-time multimodal guidance—including fluoroscopy, DSA, and videoconferencing tools (e.g., Zoom, Jitsi) increased operators’ situational awareness. Initial security protocols (utilization of VPNs) indicate an increasing focus on data integrity. 

However, the field remains in its early developmental stages. Most investigations focus on fundamental technology validation, predominantly employing phantom models or animal studies. Only two studies have progressed to prototype demonstrations in clinical environments. According to \gls{trl} guidelines~\cite{mankins1995technology}, most of the studies are in Level 3~5, the low \gls{trl} of most studies highlights a translational gap that needs to be bridged to achieve clinical validation. All included studies exhibited a high risk of bias when assessed with ROBINS-I and QUADAS-2, mainly due to heterogeneous protocols, non-standardized patient-selection criteria, and inconsistent reference standards. The lack of standardization methodologies between studies limits cross-study comparisons. 

\subsection{Limitations of this review}
A potential limitation of our review lies in its scope, which restricted itself to endovascular intervention teleoperation. Therefore, no telesurgery studies in other medical fields were examined (e.g., laparoscopic or orthopedic domains). However, our aim was to provide a precise understanding of teleoperation of endovascular interventions which is a highly specialized area with unique challenges. 
An additional limitation arises from our focus on systems evaluated under remote teleoperation conditions, defined as interventions conducted over distances exceeding \SI{100}{\meter}. While this criterion was adopted to emphasize clinically relevant long-distance teleoperation, it inherently excludes some studies that investigate endovascular robotic systems under short-range or next-room configurations. Although these studies do not primarily focus on remote scenarios, they may nevertheless report advances in actuation mechanisms, sensing modalities, control strategies, and system integration that are not captured within the scope of this review.
Another limitation is the exclusion of pre-prints and non-peer-reviewed material which may introduce publication bias. Owing to disparate publication pace between peer-reviewed journals and pre-prints, it is conceivable that clinically-orientated teams may be more inclined to publish in a peer reviewed journal compared to more robotic-orientated teams who may publish a pre-print prior to peer-review concluding.
\subsection{Limitations of Teleoperated Endovascular Interventions}
% Limitation 3
%Methodological and Clinical Translation Gaps:
\paragraph{Study Design and Methodological Shortcomings}
Currently, teleoperation of endovascular interventions still remains at an early translational stage. Alongside the high bias risk in most studies, the level of evidence for current endovascular teleoperation efficacy is low~\cite{OCEBM2011}. Indeed, only 12.5\% (2/16) of eligible studies were clinical involving humans, and these included a small number of patients. Furthermore, the inclusion criteria for these patient groups was restricted, limiting generalizability~\cite{patel_long_2019,eleid_remote_2021}. Additionally, the lack of standardized reporting frameworks for metrics is problematic for cross-study comparison, which requires comprehensive criteria. These reporting criteria should include true end-to-end latency, defined as the time delay between command transmission and robotic execution with video feedback. They should also include operator-robot platform calibration, which quantifies mismatches between operator inputs and robotic outputs. Bandwidth allocation, which prioritizes fluoroscopic video streaming over control signal transmission, or video resizing should also be considered.

The current latency threshold of \SI{250}{\milli\second} for \gls{pci} ~\cite{madder_network_2020,eleid_remote_2021} is derived from subjective operator ratings of task smoothness. However, no published work has rigorously quantified how transmission delays in fluoroscopic video streams impact objective performance metrics, such as catheter tip force or navigation time, even though such latency fluctuations can undermine procedural accuracy~\cite{madder_network_2020}. What has been demonstrated is that the size of the transmitted data exerts a substantially greater influence than the variation in transmission distance. Latency variations are also notably influenced by heterogeneous study parameters such as network load and procedural complexity, which differ across studies. These variations are further constrained by limited experimental repetitions, preventing definitive conclusions about optimal communication protocols and operational distance. Importantly, there is currently no study systematically evaluating the effect of delay on clinical outcomes.

In addition, most endovascular interventions (10/16, 63\%) focused on cardiac vessels, with only single studies for other critical interventions such as \gls{mt} in stroke management~\cite{singer_remote_2021}, which aligning with similar resutls found in a review of AI in endovascular interventions~\cite{robertshaw_artificial_2023}. This reveals a significant gap in teleoperation research for time-sensitive interventions such as \gls{mt}, where remote systems could offer substantial clinical benefits.

\paragraph{Clinical and Operational Dependencies}
In the studies reviewed, robotic endovascular procedures depend on local teams for initial vascular access before remote control is initiated~\cite{patel_long_2019}. On-site clinicians are still required to provide technical assistance (e.g., catheter insertion, device exchanges) which further diminishes the goal of fully remote intervention. Furthermore, expertise for emergency conversion to manual control also needs to be considered~\cite{serafini_exploring_2022}. While a hybrid approach (combined local and remote intervention) has facilitated straightforward stages of remote operations, robotic platforms remain unable to remotely perform complex tasks like thrombus retrieval during \gls{mt}, which local teams might not be able to perform~\cite{singer_remote_2021}. Although these dependencies are a limitation and pose practical challenges, it is plausible that ongoing advancements in robotic dexterity and automated device management may gradually reduce the necessity for continuous on-site support except in the event of an emergency conversion.

One study reported concerns regarding users who are unfamiliar with remote control interfaces, which may introduce inefficiencies during teleoperation~\cite{xu_long-distance_2024}. This highlights the need for additional research on advanced simulation-based teleoperation training programs and the importance of intuitive robotic interfaces~\cite{jackson2023comparative,nguyen2023advanced}. For robotic \gls{pci}, clinical teams have also reported apprehension regarding remote operators with challenges in establishing collaborative confidence between remote operators and on-site teams. This may stem from poorly designed systems or communication limitations, such as inadequate audio quality and the absence of non-verbal cues~\cite{serafini_exploring_2022}. Robust communication infrastructure, such as camera and audio equipment configurations, need also be further studied to ensure better communication between remote operators and on-site teams. Currently, no studies have published standardized remote-operation protocols, such as emergency checklists detailing contingency plans in case of connection failure, to facilitate safe clinical adoption and promote greater clinical team confidence.

% Limitation 2
% Technical and Sensory Limitations
\paragraph{Robotic System Limitations}
One teleoperation study identified compatibility limitations in the robot’s ability to accommodate specific guidewire sizes~\cite{legeza_impact_2022}. Compatibility limitations in the other studies were less clear although it is known that achieving broad adaptability across all guidewire and catheter type, remains a considerable challenge in endovascular robotics teleoperation~\cite{crinnion2022robotics}. 

Only three out of the eight robotic platforms included in this review provided haptic feedback. Though the studies without haptic feedback did not show a significant difference in the technical success rate, one study concluded that the lack of sensing the force can restrict the operator’s awareness of subtle device interactions\cite{singer_remote_2021}. The ability to integrate haptic feedback into remote robotic systems presents an ongoing challenge.
% While, the benefit of haptic feedback in endovascular robotics has not been proven~\cite{jackson2023comparative}, what is clear is that the ability to integrate tactile feedback into remote robotic systems presents an ongoing challenge.

% Limitation 4
\paragraph{Uncertain network and Communication Infrastructure}
The scalability of teleoperation hinges on proving network abilities. Although platforms demonstrated transcontinental capability~\cite{madder_network_2020}, real world validation of network reliability across varying geographic distances, service providers and diverse network conditions is still insufficient. It is also essential to anticipate and mitigate connection instability, which remains a critical obstacle. Moreover, the integration of high-bandwidth communication protocols is imperative for ensuring consistent image quality and minimal latency. Although studies addressed certain aspects of wireless performance, there remains a lack of comprehensive comparison of various standards (e.g., the definition of latency) under clinically pertinent settings.

\subsection{Future Directions}

\subsubsection{Endovascular Robot Technologies}

\paragraph{Enhancing Robot Ability and Autonomy}
To optimize the benefits of teleoperation for endovascular robotics, automation could be extended to device exchanges and contrast injection as well as tasks such as obtaining vascular access or performing advanced retrieval procedures like clot aspiration~\cite{legeza_preclinical_2021}. By allowing the robot to manage more routine tasks like contrast injection, it could help operators manage the entire procedure more comprehensively from a remote site, while also permitting the operator to direct procedure flow with minimal on-site staffing~\cite{robertshaw2024autonomous}. Procedures may also be quicker: experience from preclinical work indicates that automatic contrast injection, along with accessories such as remote foot-pedals, may help cut down on fluoroscopy time~\cite{legeza_preclinical_2021}. Furthermore, integrating autonomous navigation software or AI-based vessel-mapping features could reduce manual manipulation, enabling teleoperators to focus on decision-making aspects and immediate troubleshooting~\cite{robertshaw_artificial_2023}. It is also plausible that through the use of AI-based autonomous navigation software, there would be a safety fall-back mechanism for device control, if for example, the teleoperation connection was interrupted when navigating in critical areas such as the cerebral vessels~\cite{robertshaw2025reinforcement}.

\paragraph{Plug-and-play Systems for Quick Deployment}
One persistent limitation in telerobotic endovascular technologies is the pre-procedure setup required to match the robot’s coordinate frame to the specific catheter or guidewire in use. Modern biplane angiography systems already perform real-time geometric auto-calibration that converts an x-ray image into absolute space without user input~\cite{kaptein2011comparison}. The remote connection between the controller and robot also require further adjustment in different network settings. Building upon this clinically proven concept, a “plug-and-play” approach in which embedded sensors allow the robot to automatically identify the attached tool and refine distance would eliminate manual calibration steps. This may reduce the total procedure times~\cite{han_multi-device_2023}, which is pivotal for emergency procedures and impact clinical outcomes (e.g., MT in acute stroke)~\cite{fransen_time_2016}. 

\paragraph{Adaptive Network Methods}
Operating robotic systems over long distances riles on robust and latency-tolerant communication solutions and therefore may require buffering or predictive control algorithms that mitigate random delays and jitter~\cite{madder_network_2020}. For example, when transient network slowdowns occur, dynamic motion command adjustment could ensure consistent device control and prevent unintended sudden motion. Additionally, advanced compression methods for real-time, high-resolution imaging might mitigate large data transfer burdens~\cite{avgousti2016medical}. Parallel processing or partial rendering in the cloud also could reduce local computational loads, allowing an increase in frame rates for fluoroscopic or 3D image streams~\cite{li2024ai}.

\paragraph{Advanced Sensing and Imaging}
Some studies suggest that the absence of real-time haptic feedback has not impacted technical tasks~\cite{Cancelliere2022}. However, providing teleoperators with realistic haptic feedback data by integrating multi-axis force sensors and tactile simulators enables teleoperators to “feel” device–vessel interactions~\cite{zhang_2025_advanced}, which can reduce peak catheter contact forces and may lower the incidence of intimal injury~\cite{fagogenis2019autonomous}. 
Similarly, a framework that integrates 3D imaging with AI-driven algorithms, combining rotational cone-beam CT for real-time vessel segmentation, deformable registration, and automatic device tracking, may diminish dependence on flat 2D fluoroscopy, enabling the operator to navigate 3D reconstructions and expand beyond commercial biplane systems with rigid 3D overlays~\cite{siemens_artis_q,philips_vesselnavigator}. 
Incorporating new technologies like \gls{ar} displays can stream these adaptive 3D road maps to the remote teleoperator. Studies report shorter path-finding times and contrast savings when \gls{ar} guidance is used~\cite{elsakka2023virtual}. Integrating such \gls{ar} visualization  into the remote console therefore offers a practical route to further enhance situational awareness during remote procedure.

\subsubsection{Clinical Evaluation}
\paragraph{Standardization}
A meaningful comparison between studies requires harmonized test-beds, performance metrics and telecommunication specifications. Metrics requiring standardization include common clinical metrics (technical-success, complication and total procedure time) and teleoperation specific metrics (bandwidth class~\cite{ebihara2022tele}, “end-to-end” round-trip latency, jitter, and packet loss rate). 

\paragraph{Evidence Needed}
To justify adoption of teleoperated endovascular interventions, multi-center clinical trials that measure clinical outcomes such as morbidity and mortality are required for each use case. \gls{mt} and \gls{pci} are well placed as use cases as they require highly specialized centralized expertise, and outcomes are time dependent reducing the benefit of transferring patients~\cite{nelson_remote_2024}. Health economic studies focusing on cost-effectiveness are also needed as are usability studies incorporating operator learning curves, because current experimental settings provide controlled environments with reliable high-speed connectivity, which is crucial for teleoperation, broader implementation will require additional clinical, health economic and usability evidence generated in expanded geographic regions. In terms of low and middle income countries, feasibility studies could determine suitability in the first instance.

\paragraph{Standardized Curricula and Simulation Training}
Robotic-specific training modules for both remote operators and on-site teams are essential. The clinical teams must become familiar with bidirectional communication systems and robotic interfaces. Standardized guidelines for teams should address network connection failures, hardware failures, device malfunctions, and emergent conversions. 

\paragraph{Cyber-security and Ethical Considerations}
Teleoperated endovascular intervention transmits high-resolution imaging and patient data across the internet, hence robust data security and cyber-resilience are mandatory. Advanced encryption protocols including multi-factor authentication could reduce network vulnerabilities which might otherwise compromise patient privacy~\cite{serafini_exploring_2022}. The vulnerabilities of real-world robotic system require further development, including regular penetration testing, encrypted control channels, and well-defined responsibility frameworks~\cite{gordon2022protecting}. Ethical and legal aspects of teleoperation must also be carefully addressed~\cite{patel2024technical}.

\subsubsection{Wireless Communication Technologies}
\paragraph{Satellite Networks} 
The rapid development of modern telecommunication technologies offers new opportunities for addressing connectivity challenges in regions with underdeveloped telecommunication infrastructure. Traditional terrestrial infrastructure, often constrained by geographical barriers, high deployment costs, and logistical complexities, has struggled to deliver reliable connectivity in underdeveloped regions. Satellite networks, such as Starlink (Starlink Services LLC, Hawthorne, USA), Iridium (Iridium Communications Inc., McLean, USA), GuoWang (China SatNet, Beijing, China) and OneWeb (Eutelsat OneWeb, London, UK) circumvent these limitations by delivering scalable, low-latency service that reduces reliance on ground-based infrastructure. These networks can deliver connectivity across vast and remote regions, including areas with little or no terrestrial coverage. Establish connection through such satellite systems could extend teleoperated endovascular interventions to underserved areas.

\paragraph{Over-the-Air Computation}
Over-the-air computation (AirComp) is a communication technique that exploits the signal superposition property of wireless multiple-access channels to directly compute functions of distributed data during transmission \cite{csahin2023survey}. Instead of sending raw data from multiple sensors to a central processor for aggregation, AirComp enables simultaneous transmission of analogue-modulated signals, allowing the receiver to obtain a function (e.g., sum, mean) of the data “in the air.” This approach reduces latency and communication overhead, making it particularly suitable for applications requiring real-time data fusion and low-latency control, such as remote robotic systems and teleoperation. In teleoperated endovascular robotics, AirComp can facilitate fast aggregation of sensor measurements, like haptic feedback or vital signs, improving responsiveness and system efficiency without increasing bandwidth requirements.

\paragraph{Integrated Sensing and Communication (ISAC) \& Semantic Communication}
Teleoperated endovascular procedures require transmission of multiple sources of data, including high-resolution fluoroscopy images, haptic feedback data (if present), audio-visual signals (operating room conditions), patient vital signs, and potentially, instructional text. Given that both sensing and communication systems typically share common frequency bands, such as the millimetre-wave (mmWave) spectrum, this overlap can lead to spectrum congestion and performance limitations. Two complementary techniques may help to solve the data transmission problem:

\begin{itemize}
\item \gls{isac}: \gls{isac} offers a unified framework that seamlessly integrates sensing and wireless communication functionalities within a single system, leveraging a shared spectrum and potentially shared hardware resources. This configuration enables simultaneous and interference-aware data transmission~\cite{liu2020joint}. By jointly modulating radar waveforms onto the communication carrier or by designing joint signals, \gls{isac} re-uses the spectrum and enables a synergistic relationship, lets radar echoes and user data occupy the same time–frequency resources instead of being separated by guard bands~\cite{liu2022integrated}. Camera-assisted \gls{isac} has also developed recently, enabling the combination of high-definition imaging with intelligent processing techniques, thereby improving surgical accuracy and real-time responsiveness~\cite{lu2024semantic}.
In teleoperation scenarios, the same \gls{isac} link must carry bidirectional, low-latency audio plus several synchronized video streams from spatially distributed cameras, giving the remote operator full situational awareness. Commercial platforms for video-conferencing generally offer limited viewpoints. Embedding the multi-camera uplink and the haptic-feedback downlink in an \gls{isac} waveform therefore delivers spectrum efficiency.
\item Semantic communication: Semantic communication emphasizes the extraction and transmission of contextually relevant and meaningful information, reducing redundant data transmission and bandwidth demands, and minimizing latency~\cite{luo2022semantic,yang2024healthcare}. For example, a semantic communication system can selectively prioritize key video frames or essential features, such as endovascular instrument movements instead of transmitting complete video streams in full resolution~\cite{li2024video}. Similarly, haptic feedback can be streamlined by encoding only pertinent tactile information, omitting redundant signals~\cite{vittorias2009perceptual}. This targeted approach not only alleviates network congestion but enhances transmission reliability, thus improving teleoperated precision and efficacy~\cite{ebihara2022tele}.
\end{itemize}

\paragraph{Edge Intelligence}
Edge intelligence refers to the integration of real-time data processing and AI-driven decision-making at or near the end device (responder-side intervention robots). This configuration removes the need for additional cloud-based data transfer, thereby decreasing latency, conserving bandwidth, and ensuring greater data privacy~\cite{luo2021resource}. Endovascular teleoperations demand ultra-low latency to synchronize the operator’s actions with robotic counterparts. Even slight delays can undermine surgical precision, compromising patient outcomes. Only critical data is subsequently transmitted to centralized systems, optimizing bandwidth usage and ensuring timely delivery of actionable insights~\cite{lim2022realizing}. By processing and temporarily storing sensitive information locally, edge intelligence is able to reduce vulnerabilities associated with cloud-based data transfers.

\subsubsection{Sensing Technologies}

\paragraph{Multimodal and Multiband Sensing} 
Multimodal and multiband sensing strategies enhance perception by combining data from diverse sensing modalities and frequency bands~\cite{el2020review}. Multimodal sensing integrates heterogeneous data sources, such as audio, visual and haptic feedback, to provide a richer and more comprehensive understanding of the operative environment. In contrast, multiband sensing fuses data captured across different frequency bands, enabling the system to exploit complementary signal characteristics while improving robustness and adaptability. When applied to teleoperation, the integration of these sensing approaches allows sensor data from multiple modalities and frequency bands to be combined efficiently. This not only reduces signal cross-talk and leverages underutilized spectral resources but also enhances signal integrity without requiring additional cabling. As a result, it improves overall sensing performance and situational awareness for the remote operator.

\paragraph{Neuromorphic Sensing}
Neuromorphic sensing is an emerging paradigm inspired by the structure and function of biological sensory systems, particularly the human brain and nervous system~\cite{jiang2023tetrachromatic}. Unlike conventional sensors that sample at fixed rates, neuromorphic sensors are event-driven: they transmit data only when a meaningful change occurs. This approach enables ultra-low-latency signal processing while cutting bandwidth and power consumption, which is beneficial for telesurgical systems.
For example, neuromorphic vision sensors can capture rapid changes in contrast within fluoroscopic or intravascular imaging sequences, aiding precise navigation and real-time decision-making. When integrated with haptic feedback and physiological monitoring, neuromorphic sensing enhances the robot’s situational awareness and enables more intuitive teleoperation. 

\paragraph{Near-field Effects in High-Frequency Sensing}
Near-field sensing refers to the use of electromagnetic fields in the immediate vicinity of a sensor, typically within one wavelength of the signal, to capture high-resolution spatial and physiological information. Unlike far-field systems that rely on radiated waves for distant target detection, near-field sensing exploits evanescent or reactive fields, enabling fine-grained, localized measurements with reduced interference and high sensitivity \cite{cong2024near}. In the context of endovascular robots, near-field sensing can be used for precise tracking of catheter tips, detecting tissue proximity, or monitoring subtle physiological changes such as blood flow or vessel wall motion. Millimeter-wave and terahertz-based near-field sensors are especially promising due to their compact form factor and the ability of the generated waves to penetrate biological tissues at shallow depths.

\subsubsection{Other Technologies and Artificial Intelligence}
Beyond the wireless communication and sensing techniques discussed above, several emerging concepts can further strengthen teleoperated endovascular intervention. A concise overview is provided below.

% \paragraph{Digital Twin} A digital twin fuses pre-operative and intra-operative material, and physical information from endovascular robots, to build a physics-based model that evolves synchronously with the live procedure ~\cite{asciak2025digital}. This approach allows the remote operator to have a more comprehensive view of the whole procedure and allows better monitoring of the performance~\cite{shu2023twin}. One study enabled a similar digital twin method in a simulator environment~\cite{serafini_exploring_2022}, although this could be developed further.

\paragraph{Split Computation Method} Split computation is an architectural method that partitions a workload across two or more processing tiers. The computation during the teleoperation procedure can be split across various computers to lower both the transmission cost and computation time~\cite{meshram20205g}. 

\paragraph{Generative-AI Reconstruction} Generative-AI reconstruction in live fluoroscopic video transmission uses deep generative models to rebuild high-fidelity frames from bandwidth-friendly, low-resolution or sparsely coded streams~\cite{ghodrati2021temporally}. By hallucinating intermediate fluoroscopy frames, this method allows remote operators to receive low-latency video even over constrained networks and enhances situational awareness when streaming at lower frame rates.

\paragraph{Immersive Communication}
%Immersive communication augments traditional telepresence with additional signals, such as extended-reality visuals, spatial audio, and haptic feedback, which allow remote users to feel co-located in a shared 3D scene~\cite{Perez2022EmergingImmersive}. When these multimodal channels are integrated into teleoperated endovascular intervention, a surgeon wearing a \gls{ar} headset can view a stereoscopic rendering of live fluoroscopy and is able to see the operating room for better situational awareness, while a robot executes the corresponding motions at the patient’s bedside. With the use of generative-AI reconstruction, immersive communication can achieve high resolution and low latency. 
Immersive communication enhances traditional telepresence by incorporating extended-reality visuals, spatial audio, and haptic feedback, enabling remote users to feel co-located within a shared 3D environment~\cite{Perez2022EmergingImmersive}. 
When applied to teleoperated endovascular intervention, these multimodal channels enable a surgeon equipped with a \gls{ar} headset or holographic display to view stereoscopic live fluoroscopy while remaining aware of the surrounding operating room environment. Meanwhile, a robotic system executes precise actions at the patient’s bedside. The integration of generative AI for scene reconstruction further enhances the experience by enabling high-resolution, low-latency rendering, thereby improving visual realism and system responsiveness.

%\item \textbf{Distributed-AI sensing}: Distributed-AI sensing deploys multiple AI agents embedded in robotic systems, edge gateways and cloud nodes to interpret diverse sensor streams cooperatively and in real time~\cite{wen2024survey}. Each agent handles the data closest to its source for ultra-low latency insights, while aggregated intelligence across the network yields richer scene understanding and predictive analytics~\cite{wen2024survey}. This structure can potentially handle node failures, boosting precision, safety, and resilience during teleoperation procedures, which is a safeguard for rural or less-developed region deployments.

\paragraph{Distributed-AI}
Distributed AI is a system architecture in which multiple intelligent agents or computational nodes collaborate by sharing information, distributing tasks, and making collective decisions to solve complex problems. Unlike centralized AI, distributed AI exploits parallel processing and geographic decentralization, providing greater scalability, robustness, and reduced latency. In the context of telesurgery, distributed AI enables decentralized processing of sensor data, adaptive control, and real-time decision-making across interconnected robotic components and remote operators \cite{duan2022distributed}. In the clinic, individual edge devices can locally analyze portions of sensor data, alleviating computational burden on any single node and enhancing resilience to network delays or failures. This distributed approach facilitates seamless integration of multimodal sensing and communication, ultimately improving the precision, responsiveness, and safety of remote surgical procedures.

\paragraph{Trustworthy AI} Trustworthy AI is essential in surgically relevant contexts; it must be demonstrably safe, reliable, transparent, and ethically aligned while assisting with or automating surgical tasks across a network. Achieving this requires rigorous validation of algorithms, transparent decision-making processes, data-privacy protections, ongoing performance assessment, human oversight, patient-welfare safeguards, and clear clinician responsibility~\cite{ucsf_trustworthy_ai}.

\section{Conclusion}
Teleoperated robotics for endovascular procedures have proven technically viable, achieving 100\% procedural success in two small human \gls{pci} clinical trials. However, most of the supporting evidence still stems from phantom or animal studies, and the low \gls{trl}, coupled with considerable risk of bias in available studies highlight a translational gap that needs to be bridged to achieve clinical validation. Standardised reporting of performance metrics would allow for meaningful comparison across investigations. Advances in haptic feedback, AI-assisted navigation, and resilient telecommunication networks are expected to enhance precision, bolster operator confidence, and expand global accessibility. Ultimately, robust prospective evidence with harmonized methodologies will be required to establish clinical validity and facilitate widespread adoption.

%% file: main_new.bbl
% Auto-generated from references.bib for arXiv portability.
% Regenerate from references.bib if citations are changed.